\documentclass[10pt,twocolumn,letterpaper]{article}

\usepackage[table]{xcolor}
\usepackage{etoolbox}
\usepackage{pgf} 

\definecolor{high}{HTML}{00AEFF}
\definecolor{low}{HTML}{FFE000}

\usepackage{cvpr}
\usepackage{multirow}
\usepackage{booktabs}
\usepackage{pifont}
\newcommand{\cmark}{\ding{51}}%
\newcommand{\xmark}{\ding{55}}%

\usepackage[accsupp]{axessibility} 

\definecolor{cvprblue}{rgb}{0.21,0.49,0.74}
\usepackage[pagebackref,breaklinks,colorlinks,citecolor=cvprblue]{hyperref}

\title{Localization Is All You Evaluate: \\ Data Leakage in Online Mapping Datasets and How to Fix It}
\author{Adam Lilja$^{1,2}$
\quad Junsheng Fu$^{2}$
\quad Erik Stenborg$^{2}$ 
\quad Lars Hammarstrand$^{1}$
\\
\normalsize$^1$Chalmers University of Technology \hspace{0.8cm} $^2$Zenseact\\
{\tt\small \{firstname.lastname\}@\{chalmers.se, zenseact.com\}}
}

\def\oursplitname{Near}
\def\originalsplitname{Orig}

\def\lanemarkers{Divider}
\def\roadedges{Boundary}
\def\zebras{Crossing}

\begin{document}
\maketitle
\begin{abstract}
The task of online mapping is to predict a local map using current sensor observations, e.g. from lidar and camera, without relying on a pre-built map. 
State-of-the-art methods are based on supervised learning and are trained predominantly using two datasets: nuScenes and Argoverse 2. 
However, these datasets revisit the same geographic locations across training, validation, and test sets. 
Specifically, over $80$\% of nuScenes and $40$\% of Argoverse 2 validation and test samples are less than $5$ m from a training sample. 
At test time, the methods are thus evaluated more on how well they localize within a memorized implicit map built from the training data than on extrapolating to unseen locations. 
Naturally, this data leakage causes inflated performance numbers and
we propose geographically disjoint data splits to reveal the true performance in unseen environments. 
Experimental results show that methods perform considerably worse, some dropping more than $45$ mAP, when trained and evaluated on proper data splits. 
Additionally, a reassessment of prior design choices reveals diverging conclusions from those based on the original split. 
Notably, the impact of lifting methods and the support from auxiliary tasks (e.g., depth supervision) on performance appears less substantial or follows a different trajectory than previously perceived.  
\href{https://github.com/LiljaAdam/geographical-splits}{https://github.com/LiljaAdam/geographical-splits}
\vspace{-0.5cm}
\end{abstract}    
\section{Introduction}
\label{sec:intro}
A core capability for an autonomous vehicle is to estimate the road in its vicinity. 
There are two complementary approaches for this task: retrieving the information from a pre-built map using localization \cite{chalvatzaras2022survey}, and directly predicting the online map using onboard sensors like camera and lidar \cite{liao2023maptrv2}. The former, \emph{Online Map Retrieval} (OMR), assumes there exists a map over the deployment area, while the latter, \emph{Online Map Estimation} (OME) assumes no such map exists.
A pre-built map provides detailed information but also requires robust localization and continuous map updates to be useful. OME sidesteps this and instead relies solely on onboard sensors and algorithms. It is thus independent of variations in current surroundings compared to mapped data. 
The challenge with OME instead lies in generalizing to new locations, beyond the places captured in the training data.

\begin{figure}[t]
  \centering
   \includegraphics[width=\columnwidth, trim={110mm, 0mm, 90mm, 0}, clip]{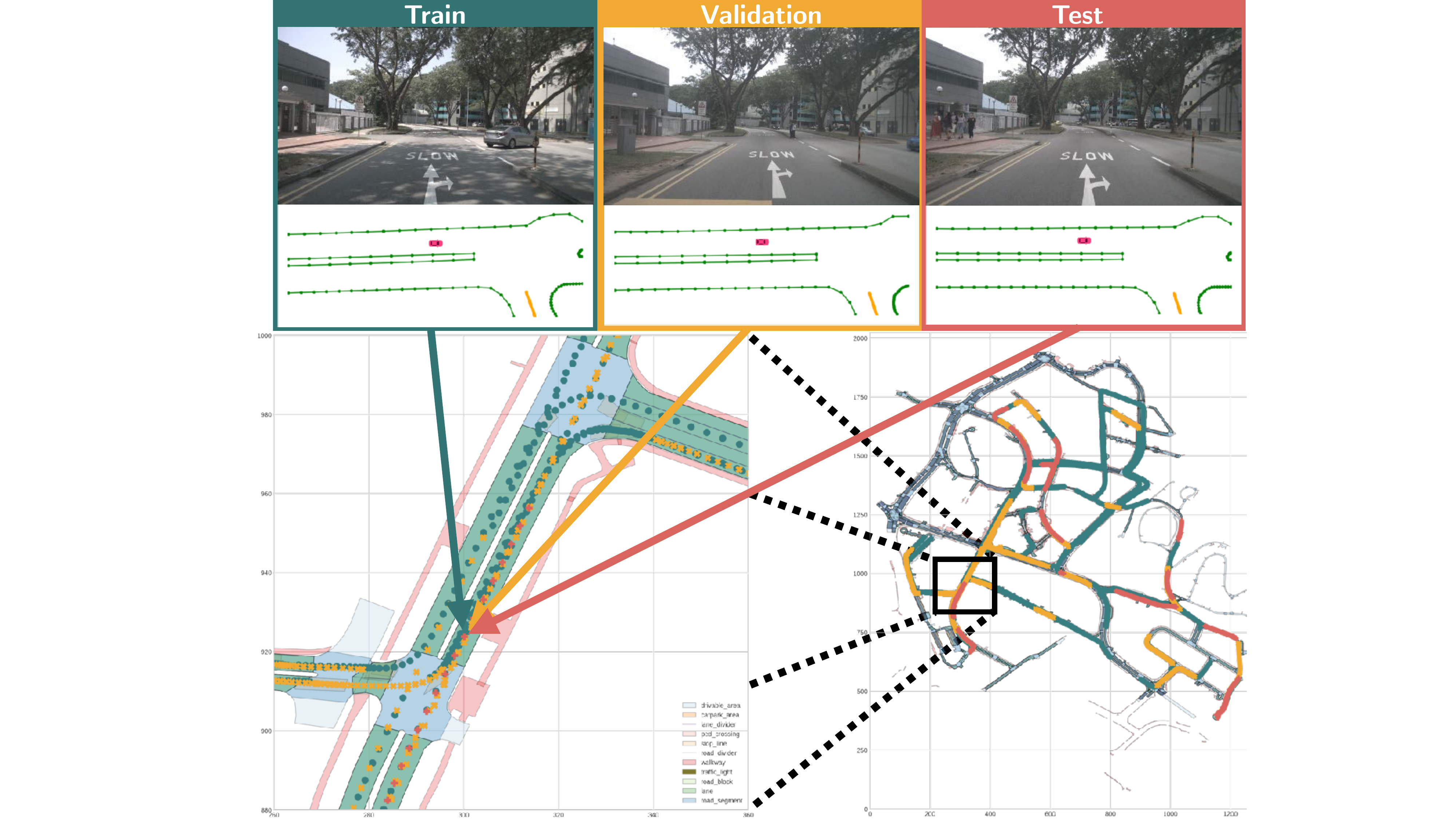}
   \caption{Example of substantial geographical overlap between train, val, and test sets for in nuScenes'. Green circle, Orange cross, and Red plus are training, validation, and test samples.}
   \vspace{-0.5cm}
   \label{fig:qual-example-geo-overlap}
\end{figure}
\begin{figure*}[t]
  \centering
   \includegraphics[width=\linewidth, trim={0mm, 90mm, 0mm, 80mm}, clip]{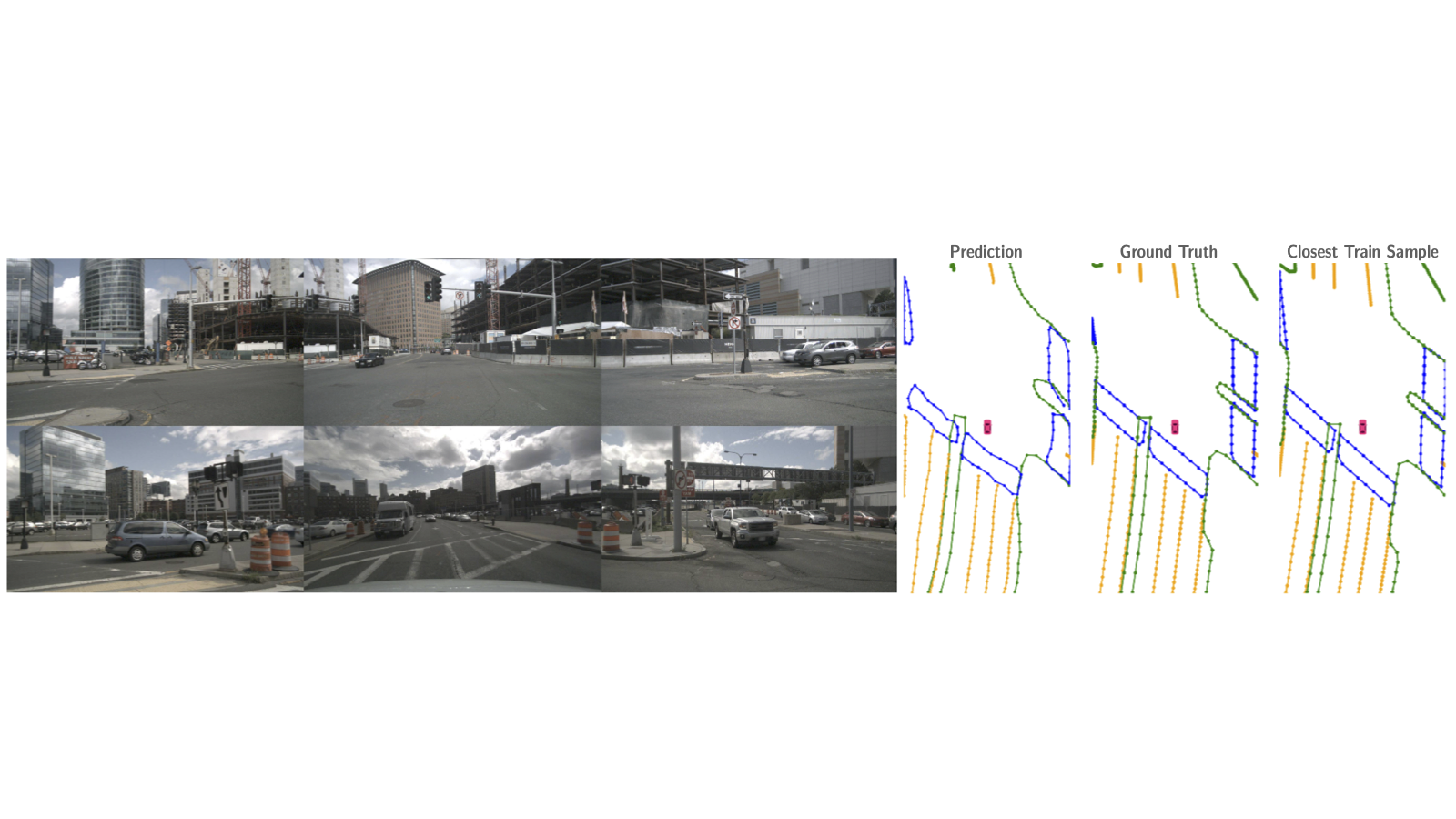}
   \caption{Example from nuScenes where input images, predictions, and ground truth for a test sample are displayed together with the ground truth of the closest train sample. Lane dividers, road boundaries, and pedestrian crossings are visualized in orange, green, and blue.}
   \vspace{-15pt}
   \label{fig:maptr-pred-gt-traingt}
\end{figure*}

The current state-of-the-art methods for online mapping are based on supervised learning. While there exist large public datasets \cite{Geiger2013IJRR, waymo_2020, yu2020bdd100k, alibeigi2023zenseact, nuscenes_cvpr_2020, argoverse1_2019, argoverse2_2021} that support training of perception and planning models for many of the crucial tasks of an autonomous vehicle, only a few of these provide the HD maps needed to train online mapping models, see \cref{tab:dataset}. 
Moreover, as these datasets are mainly constructed to support object detection, object tracking, and motion forecasting tasks, we argue that they, in their original form, are not ideal for training online mapping models for two reasons: 
(1) The training, validation, and test sets are constructed by splitting the data temporally. 
This is an easy way to ensure that, \eg, the same vehicle is not present in the same position in the training and test sets yielding fair evaluation of object detection methods. 
However, since the areas where these datasets are collected are relatively small, the same areas are revisited multiple times. 
Failing to account for this when dividing the data, results in significant geographical \emph{inter-set} overlap between the training, validation and test sets as \cref{fig:qual-example-geo-overlap} exemplifies. 
\cref{fig:maptr-pred-gt-traingt} further visualizes the input images, prediction, ground truth, and the closest training sample of a test sample for another example. 
The test sample is very close to the nearest training sample, enabling a method to score well on that test sample by memorizing the image-to-map connection from the training sample, and recalling it at test time. This connection can be created using any features in the images such as buildings and traffic signs.
This would rather resemble localizing in a pre-built map and presenting it at test time (OMR), than the intended task of online mapping (OME).
(2) As each data sample is collected from a data sequence under normal driving, there is substantial \emph{intra-set} overlap/correlation between the data samples within the sets. This correlation is especially evident near intersections where the vehicle is standing still or driving slowly. Essentially, the ground truth HD maps for close training samples are only slight transformations of the same information. Both these aspects violate the independent and identically distributed data assumption fundamental in training machine learning models.  

In this paper, we show that there is a significant geographical overlap in two of the most commonly used datasets for online mapping, nuScenes \cite{nuscenes_cvpr_2020} and Argoverse 2 \cite{argoverse2_2021}. 
Furthermore, we show that this overlap causes severe inflation in the reported performance of the state-of-the-art methods and, more critically, results in poorer generalizability than initially perceived.

To support future research in OME we provide geographically disjoint splits for both nuScenes and Argoverse 2. 
We provide splits under two slightly different problem settings,  \emph{Near Extrapolation} where we assume we have training data from the same neighbourhood/city and \emph{Far Extrapolation} where training and test data are from separate cities. 
The former is an easier setting and can be viewed as a proper substitution for the original splits while the latter enables the exploration of the more relevant question; how well do these methods generalize to new environments with larger distribution shifts?

We re-evaluate state-of-the-art methods trained on these splits to give a more representative view of their performance. 
Additionally, we perform more detailed experiments to investigate how the large overlap has affected conclusions regarding important algorithmic choices.

\section{Related Work}
\label{sec:related-work}
This section introduces the online mapping setting (OME), highlights key methodologies and datasets used for training and evaluation. 
While the OME field is relatively recent, the utilization of machine learning in processing geospatial data (GeoML) has a longer history. 
Hence, we also provide a brief overview of GeoML, noting parallels between its challenges and those encountered in OME.

\subsection{Online Mapping}
The current online mapping methods are either segmentation-based \cite{philion2020lift, roddick2020predicting, xie2022m, li2022bevformer, liu2022bevfusion, dong2023superfusion, li2022hdmapnet, chen2022efficient, hu2022stp3, zhang2022beverse, Peng_2023_WACV} or vector-based \cite{li2022hdmapnet, liu2023vectormapnet, liao2023maptr, shin2023instagram, liao2023maptrv2}. The main difference lies in how the online map is represented. In segmentation-based maps, the aim is to predict a rasterized grid where each cell is classified as, e.g., empty, lane marking or road edge. For vector-based methods, the predicted map is described by a set of objects with a given class and the geometry is described by a vector of point coordinates, \ie, a polyline.

Both categories universally adopt a core technique known as \emph{lifting}. This entails converting image features from perspective view (PV) to Bird's Eye View (BEV) features, from which the map is predicted. The primary challenge for these methods lies in accurately mapping features from the perspective view to their corresponding locations in BEV due to the absence of depth information in images.
Various lifting methods have been proposed, broadly categorized as either \emph{pulling} the features to BEV from PV, or \emph{pushing} the features from PV to BEV.

In essence, \emph{pulling} methods retrieve features from PV based on dense queries in BEV \cite{chen2022efficient, li2022bevformer, zhou2022crossview, Peng_2023_WACV}. 
A straightforward approach is the Inverse Perspective Mapping (IPM) \cite{mallot1991inverse} and involves projecting predefined points in BEV into PV using camera parameters and interpolating features from these projected positions.
Alternatively, methods like Geometry-Guided Kernel Transformer (GKT) \cite{chen2022efficient} and BEVFormer \cite{li2022bevformer} use a combination of geometry and attention mechanisms to pull features to BEV-space efficiently.  
In contrast, Cross-View Transformer (CVT) \cite{zhou2022crossview} pulls features without an intermediate BEV representation using cross-attention with a canonical form of all camera views. 

The \emph{pushing} methods used in, \eg \cite{philion2020lift,roddick2020predicting,xie2022m,liu2022bevfusion, dong2023superfusion, li2022hdmapnet}, specialize in learning how to map PV features to BEV. Among them, depth-based approaches aim to learn the depth distribution for each image pixel to project PV features accurately. For instance, LSS \cite{philion2020lift} tries to learn a categorical depth distribution for each pixel and use it to weigh how much the corresponding PV feature should influence the corresponding BEV-cell. 
Pyramid Occupancy Network (PON) \cite{roddick2020predicting} uses a multi-scale dense transformer for low-resolution BEV projection, employing deconvolutions for upsampling predictions, whereas 
HDMapNet \cite{li2022hdmapnet} learns BEV projection through a Multilayer Perceptron (MLP). 

These lifting methods have been successfully adapted to a segmentation head for online mapping. Also the vector-based approach, introduced in HDMapNet \cite{li2022hdmapnet} and further developed in works such as \cite{liu2023vectormapnet, liao2023maptr, shin2023instagram, liao2023maptrv2}, have shown great promise by utilizing network heads inspired by the object-detection community.
While HDMapNet uses a handcrafted post-processing step, subsequent methods are instead end-to-end trainable. 
For example, VectorMapNet \cite{liu2023vectormapnet} uses IPM \cite{mallot1991inverse} for lifting image features to BEV from which a transformer decoder predicts coarse object representations. These are then refined in a joint Autoregressive Transformer (ART) that attends the coarse prediction and all BEV features. 
MapTR \cite{liao2023maptr} utilizes GKT \cite{chen2022efficient} for lifting and a DETR-like \cite{carion2020endtoend} transformer decoder for predicting the objects. They use deformable attention to attend BEV features with hierarchical queries to predict a collection of objects defined by a set of points.  
MapTRv2 \cite{liao2023maptrv2} builds on its predecessor, but uses LSS\cite{philion2020lift} for lifting and adds PV depth estimation and segmentation in both PV and BEV as auxiliary supervision. 
Lastly, StreamMapNet \cite{yuan2023streammapnet} uses BEVFormer-lifting, multi-point attention, and temporal information fusion.

All these methods, except \cite{roddick2020predicting, yuan2023streammapnet}, are primarily evaluated on the original nuScenes and Argoverse 2 splits with considerable inter-set overlap. The validity of conclusions drawn regarding their performance on online mapping tasks is thus severely limited. 
To give a fairer view of their performance on the intended problem setting, we re-evaluate them on our proposed splits and analyze the results. 

\subsection{Online Mapping Datasets}
\label{subsec:online-mapping-datasets}
\begin{table}
    \centering
    \renewcommand{\arraystretch}{0.8} 
    \setlength{\tabcolsep}{2pt} 
    \scalebox{0.8}{%
    \begin{tabular}{@{}ll cccccc@{}}
        \hline
        \toprule 
        \multirow{2}{*}{Dataset} & \multirow{2}{*}{Split} & \multirow{2}{*}{Source} & \multirow{2}{*}{\begin{tabular}{c}Main Map \\ Purpose\end{tabular}} & \multicolumn{3}{c}{\#Samples} & \multirow{2}{*}{\begin{tabular}{c}Geo. \\ Split\end{tabular}} \\
        \cmidrule(lr){5-7} 
        & & & & Train & Val & Test & \\ 
        \midrule
        nuScenes \cite{nuscenes_cvpr_2020}     & Original       & nuSc  & OD/MF &$28$k&$6$k&$6$k     &\xmark \\
        Argoverse 1 \cite{argoverse1_2019}     & Original       & argo1 & OD/MF &$39$k&$15$k&$13$k   &\cmark \\
        Argoverse 2 \cite{argoverse2_2021}     & Original       & argo2 & OD/MF &$110$k&$24$k&$24$k  &\xmark \\
        Waymo \cite{waymo_2020}                & Original       & way   & OD/MF &$122$k&$30$k&$40$k  &\xmark \\
        \midrule
        nuScenes \cite{nuscenes_cvpr_2020}    & \textbf{\oursplitname}        & nuSc  & OM &$28$k&$6$k&$6$k       & \cmark\\
        Argoverse 2 \cite{argoverse2_2021}    & \textbf{\oursplitname}        & argo2 & OM &$110$k&$24$k&$24$k    & \cmark\\
        \midrule 
        nuScenes \cite{nuscenes_cvpr_2020}    & \textbf{Far-A}   & nuSc  & OM &$30$k&$9$k&-      & \cmark\\
        nuScenes \cite{nuscenes_cvpr_2020}    & \textbf{Far-B}   & nuSc  & OM &$31$k&$9$k&-      & \cmark\\
        Argoverse 2 \cite{argoverse2_2021}    & \textbf{Far-A}   & argo2 & OM &$110$k&$46$k&-    & \cmark\\
        Argoverse 2 \cite{argoverse2_2021}    & \textbf{Far-B}   & argo2 & OM &$101$k&$55$k&-    & \cmark\\
        Argoverse 2 \cite{argoverse2_2021}    & \textbf{Far-C}   & argo2 & OM &$101$k&$55$k&-    & \cmark\\
        \bottomrule 
    \end{tabular}}
    \caption{Datasets used for online mapping. The proposed splits are shown in bold. OD = object detection, MF = motion forecasting, OM = online mapping. }
    \label{tab:dataset}
    \vspace{-0.2cm}
\end{table}


\begin{table}
    \centering
    \renewcommand{\arraystretch}{0.8} 
    \setlength{\tabcolsep}{2pt} 
    \scalebox{0.8}{%
    \begin{tabular}{l cc cc}
    \hline 
    \toprule 
    \multirow{2}{*}{Split} & \multicolumn{2}{c}{nuScenes} & \multicolumn{2}{c}{Argoverse 2} \\ \cmidrule(lr){2-3} \cmidrule(lr){4-5}
     & Val & Test & Val & Test \\ \midrule
    Orig.  & $79.4$\% & $85.5$\% & $45.0$\% & $41.9$\% \\ 
    Near   & $0.9$\%  & $1.1$\%  & $0.0$\% & $0.0$\% \\
    \bottomrule 
    \end{tabular}}
    \caption{Ratios of validation and test samples within $5$ m of training samples. The Near Extrapolation split has negligible overlap compared to the Original (Orig.) split for both datasets. }
    \vspace{-0.5cm}
    \label{tab:overlap-ratios}
\end{table}

A summary of datasets used for online mapping is provided in \cref{tab:dataset}. Three original datasets provide the HD-maps required to enable the training of online mapping models, Argoverse 1 and 2 \cite{argoverse1_2019, argoverse2_2021}, nuScenes \cite{nuscenes_cvpr_2020} and Waymo \cite{waymo_2020}. 
All these primary datasets are mainly intended for object detection and motion forecasting tasks, and, in addition to supplying HD maps, these datasets provide rich annotations for dynamic objects. 
The predefined data splits provide fair and consistent evaluations across studies, but were originally designed for dynamic object perception rather than online mapping. They are temporally divided to prevent sample overlap across sets within a sequence, but do not ensure geographic separation. Despite this, nuScenes \cite{nuscenes_cvpr_2020} and Argoverse 2 \cite{argoverse2_2021} are widely used for training online mapping models and have become the \emph{de facto} standard.
For example, online mapping methods using nuScenes include \cite{li2022hdmapnet, chen2022efficient, zhou2022crossview, philion2020lift, roddick2020predicting, xie2022m,jiang2023polarformer, li2022bevformer, liu2022bevfusion, dong2023superfusion, saha2022translating, hu2022stp3,Qin_2023_ICCV, zhang2022beverse, Peng_2023_WACV, can2021structured} and Argoverse 2 is used in \cite{liu2023vectormapnet, shin2023instagram, liao2023maptr, liao2023maptrv2}. 

The nuScenes dataset contains $1\,000$ driving sequences collected in two cities (Boston and Singapore) with an area coverage of about $5$ km$^2$ \cite{waymo_2020} and captures different types of city roads as well as containing diverse weather and illumination conditions. 
In total, the sequences consist of $40,000$ key-frame samples at a rate of $2$ Hz, accompanied by object annotations. Additional sensor data is present between these key-frame samples, albeit without any object annotation. 
Online mapping methods typically adhere to the convention established by object detection methods, utilizing only key-frame samples for training.
Furthermore, these samples are closely together across the different sets as the same geographical position is re-visited multiple times. \cref{fig:qual-example-geo-overlap} illustrates a single example, where the proximity of validation and test samples to training samples is evident. Further analysis, as  \cref{tab:overlap-ratios} depicts, reveals that approximately $80\%$ and $85\%$ of validation and test samples, respectively, are within $5$ meters of a sample used during training.

The other prominently used dataset, Argoverse 2, is an extension of the 2019 version Argoverse \cite{argoverse1_2019}. In contrast to its predecessor, Argoverse 2 is not geographically split, but is much larger and collected from 6 U.S. cities with an area coverage of $17$ km$^2$ \cite{alibeigi2023zenseact}. It comprises $1\,000$ annotated driving sequences, which are on average $15$ s long and annotated at $10$ Hz. 
Each sample offers observations from similar sensors as nuScenes, albeit in a slightly different configuration, and the provided HD maps are in 3D, but with a focus on drivable area, road boundaries, lane dividers, and pedestrian crossings. 
By inspecting Argoverse 2, one can see that it also suffers from a considerable level of inter-set geographical overlap. 
\cref{tab:overlap-ratios} shows that approximately $45\%$ and $42\%$ of the validation and test samples are within a $5$ m range of the closest train sample. 

\begin{figure}
    \centering
    \includegraphics[width=0.8\linewidth]{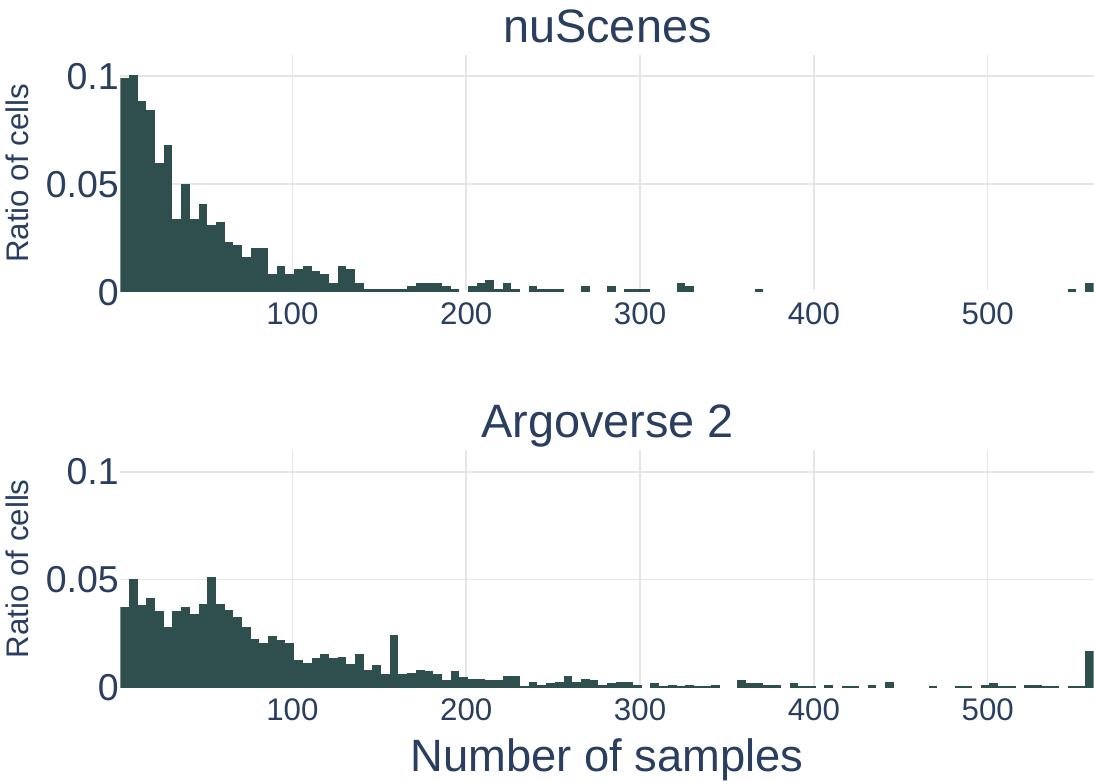}
    \caption{Number of samples within a cell with length $60$ m. Argoverse 2 has higher \emph{intra-set} sample density.} 
    \vspace{-10pt}
    \label{fig:hist_row_num_samples_per_cell}
\end{figure}

Comparing nuScenes and Argoverse 2, we note that the inter-set sample overlap is larger for nuScenes, and that Argoverse 2 boasts a larger number of samples and a more extensive coverage area. 
Another difference arises from Argoverse 2 being more densely sampled, resulting in less spatial variation and a higher intra-set sample density. 
The discrepancy is highlighted in \cref{fig:hist_row_num_samples_per_cell}, where samples are discretized into cells with a side length of $60$ m, the typical evaluation range for most online mapping methods, \eg \cite{li2022hdmapnet, liu2023vectormapnet, liao2023maptr, shin2023instagram, liao2023maptrv2}. 
The distribution of non-empty cells over the number of samples they contain has more probability mass with the higher counts for Argoverse 2. 
Despite Argoverse 2 having nearly four times the number of samples than nuScenes, the number of non-empty cells is only $1.7$ times higher. 

\subsection{Machine Learning on Geospatial Data}
Utilizing machine learning for geospatial applications has been a longstanding challenge. 
The choice of geographic regions for training and evaluation significantly impacts the evaluation outcomes and, as highlighted in \cite{eval_challenges_long}, it is crucial to select regions that mirror the desired goals for an accurate assessment.
Neglecting geographic considerations in evaluating these models can thus yield inadequate and inaccurate assessments of their performance. 
This issue, characterized by geographic overlap between training and evaluation data, has been observed in various domains, including satellite image modeling \cite{rs15153793}, ecological mapping \cite{eco_mapping_long}, medical imaging \cite{Tampu_2022}, and 3D object detection \cite{simonelli2021missing}.

In \cite{cross_val_strats_long}, the authors discuss how deviating from the assumption of independent training and evaluation data can lead to favouring complex models. 
They propose a solution known as block cross-validation, where data is strategically split geographically to mitigate these concerns. 
Similarly, \cite{geo_ai} advocates for spatially partitioning the data instead of random allocation to bolster the independence between training and validation datasets used for cross-validation.

In the online mapping community, \cite{roddick2020predicting} acknowledged the need for a geographically divided train and test split when evaluating their method (PON) on nuScenes. 
They coarsely partition the sets along large, visually similar, neighbourhoods, limiting the diversity within each set. 
Furthermore, their proposed geographical split does not take the validation set into account and discards samples from the original test set reducing the total size of the dataset. 
Another approach is to divide training and evaluation according to the different cities as explored for nuScenes in \cite{Qin_2023_ICCV}. 
Although being a valid split set, the distribution shift is large, and significantly increases the difficulty of the problem. Since only one data split is proposed, cross-validation cannot be performed.
For Argoverse 2, \cite{yuan2023streammapnet} proposes a split that divides the training and validation samples according to geographical positions, but does not allocate a geographically separate testing set. 
A separate test set representing unseen data is crucial for unbiased evaluation; hyperparameters should not be optimized specifically for this set. Hence, the need for robust data partitioning persists.

\section{Geographically Disjoint Splits}
In the online mapping setting, the difficulty level varies depending on assumptions about the geographic distribution shift between training and target data, ranging from near to far extrapolation.
Near extrapolation assesses performance within the same cities as the training data, while far extrapolation evaluates generalization to cities beyond those in the training set. 
We believe that addressing the latter scenario, which poses a greater difficulty, should be the primary long-term target of the OME field.

\subsection{Near Extrapolation}
\label{sec:new-splits}
We created balanced geographically disjoint sets for training, validation, and testing in nuScenes and Argoverse 2 by partitioning data based on sample locations, reducing inter-set overlap. 
We employ the original splits' testing data as the map is provided for those samples. 
Split proportions are $70$\%, $15$\%, and $15$\% for training, validation, and test sets.
To preserve diversity in zone classes (in the urban planning sense, e.g., residential, commercial, and industrial) while maintaining the frequency of road object classes, weather conditions, and time of day, fine-grained and thorough partitioning is performed. 
Our splits ensure proportional representation of samples from various criteria in each set, mirroring the full dataset's distribution and ensuring representation from all cities. 
Regions were defined manually based on map attributes, and splits are visualized in \cref{sec:additional-data-attributes} and available on the project webpage.

While partitioning the data, we did not account for the intra-set overlap (\cref{subsec:online-mapping-datasets}). We do, however, acknowledge its potential importance and believe it warrants consideration in the utilization of these datasets. The specifics of how to address this concern and the associated implications are left for future research. 
We realize that the use of original test data has implications for multi-task networks (e.g. also performing object detection) and have ensured that balance remains when removing the original test data. 
The number of samples in each set can be seen in \cref{tab:dataset}.
\vspace{-12pt}
\paragraph{nuScenes}
\label{sec:new-splits-nusc}
\cref{tab:overlap-ratios} displays the ratio of validation and test samples located closer than $5$ m of a training sample for the Near Extrapolation split. Our suggested splits show only minimal overlap. 
The remaining overlap is due to the samples close to cut-off regions between sets. 
To see the effects of these samples we conduct, in \cref{sec:overlapping-maps}, experiments where these samples have been filtered out and note that their impact is negligible. 
Further, we show that the weather conditions and time of day are equally distributed through the sets in our Near Extrapolation splits.
\vspace{-12pt}
\paragraph{Argoverse 2}
\label{sec:new-splits-argo}
As \cref{tab:overlap-ratios} presents, no validation or test samples lie within $5$ m of a training sample for the proposed split. 
Attributes concerning the conditions associated with the different sequences are not available for Argoverse 2 making it hard to do a quantitative analysis. Our main focus is thus to give a balanced geographically separated split, partitioning areas with similar zone classes equally in the different sets. 
\cref{sec:additional-data-attributes} illustrates the distribution of the number of samples in each city as well as highlights the diversity of geographical distribution.

\subsection{Far Extrapolation}
Far Extrapolation through city-wise data splitting introduces a greater distribution shift between training and evaluation. 
A subset of cities from each dataset is designated for training, while the remainder is allocated for evaluation. 
Notably, there is an uneven city-wise sample distribution in both datasets, with, for instance, Boston containing $55$\% of samples in nuScenes, and Miami and Pittsburgh each constituting $35$\% of the Argoverse 2 data. 
To mitigate this imbalance, cities are grouped to achieve approximately equal-sized folds, with the training set comprising $70$\% of the data. 
Refer to \cref{tab:citywise-splits} for the proposed city-wise folds in both nuScenes and Argoverse 2. 
Given the varied attributes of each city, the method's performance is sensitive to the composition of training and validation sets. 
As such, these city-wise folds ought to be utilized for cross-validation, where the average performance across different folds serves as the performance measure.

\begin{table}
    \centering
    \renewcommand{\arraystretch}{0.8} 
    \setlength{\tabcolsep}{1pt} 
    \scalebox{0.8}{%
    \begin{tabular}{c cc ccc}
    \hline 
    \toprule 
    & \multicolumn{2}{c}{nuScenes} & \multicolumn{3}{c}{Argoverse 2} \\ \cmidrule(lr){2-3} \cmidrule(lr){4-6}
    Set &                       A            &      B           & A             & B             & C                 \\ \cmidrule(lr){2-2} \cmidrule(lr){3-3} \cmidrule(lr){4-4} \cmidrule(lr){5-5} \cmidrule(lr){6-6}
    \multirow{3}{*}{Train} & Boston,         & Boston,          & Miami,        & Miami,        & Pittsburgh,       \\
                           & Onenorth        & Queenstown,      & Pittsburgh    & Rest          & Rest              \\
                           &                 & Holland Village  &               &               &                   \\ 
                           \\
    \multirow{2}{*}{Val}   & Queenstown,     & Onenorth         & Rest          & Pittsburgh    & Miami             \\
                           & Holland Village &                  &               &               &                   \\
    \bottomrule 
    \end{tabular}}
    \caption{Far Extrapolation splits for nuScenes and Argoverse 2 where the folds are approximately similarly sized.}
    \vspace{-0.5cm}
    \label{tab:citywise-splits}
\end{table}
\section{Experiments}
\label{sec:experiments}
\def\aoldlm{{$38.0$} & {$52.2$}}
\def\aoldre{{$37.2$} & {$46.0$}}
\def\aoldpc{{$24.2$} & {$28.9$}}
\def\aoldmap{{$33.1$} & {$42.4$}}

\def\boldlm{{$10.6$} & {$12.1$}}
\def\boldre{{$13.7$} & {$20.1$}}
\def\boldpc{{$5.1$} & {$1.0$}}
\def\boldmap{{$9.8$} & {$11.1$}}

\def\maptrfiftyoglm{{$51.0$} & {$65.8$}}
\def\maptrfiftyogre{{$52.6$} & {$60.5$}}
\def\maptrfiftyogpc{{$43.0$} & {$53.3$}}
\def\maptrfiftyogmap{{$48.8$} & {$59.9$}}

\def\maptrfiftynewlm{{$16.0$} & {$19.9$}}
\def\maptrfiftynewre{{$26.7$} & {$33.3$}}
\def\maptrfiftynewpc{{$14.4$} & {$5.9$}}
\def\maptrfiftynewmap{{$19.0$} & {$19.7$}}

\def\doldlm{{$48.9$} & {$47.9$}}
\def\doldre{{$40.9$} & {$63.8$}}
\def\doldpc{{$39.8$} & {$52.8$}}
\def\doldmap{{$43.2$} & {$54.8$}}

\def\eoldlm{{$13.5$} & {$17.3$}}
\def\eoldre{{$14.9$} & {$21.6$}}
\def\eoldpc{{$13.7$} & {$15.7$}}
\def\eoldmap{{$14.0$} & {$18.2$}}

\def\foldlm{{$13.9$} & {$17.9$}}
\def\foldre{{$14.0$} & {$20.8$}}
\def\foldpc{{$15.6$} & {$16.2$}}
\def\foldmap{{$14.5$} & {$18.3$}}

\def\goldlm{{$61.8$} & {$77.3$}}
\def\goldre{{$63.7$} & {$70.9$}}
\def\goldpc{{$59.1$} & {$69.8$}}
\def\goldmap{{$61.5$} & {$72.7$}}

\def\holdlm{{$20.9$} & {$23.4$}}
\def\holdre{{$32.6$} & {$40.5$}}
\def\holdpc{{$26.5$} & {$14.8$}}
\def\holdmap{{$26.7$} & {$26.2$}}

\def\ioldlm{{$20.7$} & {$23.4$}}
\def\ioldre{{$32.5$} & {$39.1$}}
\def\ioldpc{{$26.4$} & {$14.9$}}
\def\ioldmap{{$26.5$} & {$25.8$}}

\def\joldlm{{$54.9$} & {$70.5$}}
\def\joldre{{$55.1$} & {$63.9$}}
\def\joldpc{{$51.9$} & {$63.1$}}
\def\joldmap{{$54.0$} & {$65.8$}}

\def\koldlm{{$15.1$} & {$18.0$}}
\def\koldre{{$27.4$} & {$35.2$}}
\def\koldpc{{$17.4$} & {$7.0$}}
\def\koldmap{{$20.0$} & {$20.1$}}

\def\loldlm{{$15.0$} & {$18.0$}}
\def\loldre{{$27.2$} & {$35.2$}}
\def\loldpc{{$17.4$} & {$7.0$}}
\def\loldmap{{$19.9$} & {$20.1$}}

\def\moldlm{{$51.9$} & {$46.8$}}
\def\moldre{{$42.1$} & {$40.7$}}
\def\moldpc{{$38.0$} & {$38.7$}}
\def\moldmap{{$44.0$} & {$42.0$}}

\def\noldlm{{$39.8$} & {$35.0$}}
\def\noldre{{$31.5$} & {$32.4$}}
\def\noldpc{{$26.8$} & {$31.3$}}
\def\noldmap{{$32.7$} & {$32.9$}}

\def\ooldlm{{$71.7$} & {$68.9$}}
\def\ooldre{{$67.0$} & {$63.8$}}
\def\ooldpc{{$64.5$} & {$63.1$}}
\def\ooldmap{{$67.7$} & {$65.3$}}

\def\poldlm{{$58.4$} & {$56.6$}}
\def\poldre{{$51.3$} & {$53.5$}}
\def\poldpc{{$49.7$} & {$55.6$}}
\def\poldmap{{$53.1$} & {$55.2$}}

\def\qoldlm{{$68.7$} & {$66.0$}}
\def\qoldre{{$64.3$} & {$61.7$}}
\def\qoldpc{{$59.6$} & {$58.7$}}
\def\qoldmap{{$64.2$} & {$62.1$}}

\def\roldlm{{$56.2$} & {$55.0$}}
\def\roldre{{$47.8$} & {$51.0$}}
\def\roldpc{{$46.2$} & {$51.8$}}
\def\roldmap{{$50.1$} & {$52.6$}}

\def\myendcolum{15}
\begin{table*}
\centering
\renewcommand{\arraystretch}{0.8} 
\setlength{\tabcolsep}{2pt} 
\scalebox{0.8}{%
\begin{tabular}{lcccccc cc  cc  cc  cc}
\hline 
\toprule 
\multirow{10}{*}{\parbox[t]{2mm}{\multirow{10}{*}{\rotatebox[origin=c]{90}{nuScenes}}}} & \multirow{2}{*}{Model} & \multirow{2}{*}{Sensor} & \multirow{2}{*}{Backbone} & \multirow{2}{*}{Lifting} & \multirow{2}{*}{Decoder} & \multirow{2}{*}{Split} & \multicolumn{2}{c}{\lanemarkers} & \multicolumn{2}{c}{\roadedges} & \multicolumn{2}{c}{\zebras}& \multicolumn{2}{c}{Mean} \\ \cmidrule(lr){8-9} \cmidrule(lr){10-11} \cmidrule(lr){12-13} \cmidrule(lr){14-\myendcolum}
& & & & & & & Val & Test & Val & Test & Val & Test & Val & Test \\ \cmidrule{1-\myendcolum} 
& \multirow{2}{*}{VectorMapNet \cite{liu2023vectormapnet}} & \multirow{2}{*}{Camera} & \multirow{2}{*}{Resnet50} & \multirow{2}{*}{IPM} & \multirow{2}{*}{ART}  & \originalsplitname         & \doldlm & \doldre & \doldpc & \doldmap \\  
& & & & & & \oursplitname                                                                                                                          & \eoldlm & \eoldre & \eoldpc & \eoldmap \\  \cmidrule{7-\myendcolum} 
& \multirow{2}{*}{MapTR \cite{liao2023maptr}} & \multirow{2}{*}{Camera} & \multirow{2}{*}{Resnet18} & \multirow{2}{*}{GKT} & \multirow{2}{*}{DETR} & \originalsplitname                & \aoldlm & \aoldre & \aoldpc & \aoldmap \\  
& & & & & & \oursplitname                                                                                                                          & \boldlm & \boldre & \boldpc & \boldmap \\  \cmidrule{7-\myendcolum} 
& \multirow{2}{*}{MapTR \cite{liao2023maptr}} & \multirow{2}{*}{Camera} & \multirow{2}{*}{Resnet50} & \multirow{2}{*}{GKT} & \multirow{2}{*}{DETR} & \originalsplitname                & \maptrfiftyoglm & \maptrfiftyogre & \maptrfiftyogpc & \maptrfiftyogmap \\  
& & & & & & \oursplitname                                                                                                                          & \maptrfiftynewlm & \maptrfiftynewre & \maptrfiftynewpc & \maptrfiftynewmap \\  \cmidrule{7-\myendcolum} 
& \multirow{2}{*}{MapTRv2 \cite{liao2023maptrv2}} & \multirow{2}{*}{Camera} & \multirow{2}{*}{Resnet50} & \multirow{2}{*}{LSS} & \multirow{2}{*}{DETR} & \originalsplitname              & \goldlm & \goldre & \goldpc & \goldmap \\  
& & & & & & \oursplitname                                                                                                                          & \holdlm & \holdre & \holdpc & \holdmap \\  \cmidrule{7-\myendcolum} 
& \multirow{2}{*}{MapTRv2 \cite{liao2023maptrv2}} & Camera & \multirow{2}{*}{\begin{tabular}{c}Resnet50\\ SECOND\end{tabular}} & \multirow{2}{*}{LSS} & \multirow{2}{*}{DETR} & \originalsplitname        & \joldlm & \joldre & \joldpc & \joldmap \\ 
& & Lidar & & & & \oursplitname                                                                                                                          & \koldlm & \koldre & \koldpc & \koldmap \\  \cmidrule{7-\myendcolum} 
& \multirow{2}{*}{StreamMapNet \cite{yuan2023streammapnet}} & \multirow{2}{*}{Camera} & \multirow{2}{*}{Resnet50} & \multirow{2}{*}{BEVFormer} & \multirow{2}{*}{DETR} & \originalsplitname        & $64.5$ & $79.3$ & $62.3$ & $69.5$ & $61.0$ & $74.5$ & $62.6$ & $74.4$ \\ 
& & & & & & \oursplitname                                                                                                                          & $23.0$ & $22.6$ & $29.5$ & $35.2$ & $25.8$ & $26.0$ & $26.1$ &  $27.9$ \\  \cmidrule{1-\myendcolum} 
%
\multirow{6}{*}{\parbox[t]{2mm}{\multirow{6}{*}{\rotatebox[origin=c]{90}{Argoverse 2}}}}
& \multirow{2}{*}{\begin{tabular}{c}VectorMapNet \cite{liu2023vectormapnet}\\ 2D\end{tabular}} & \multirow{2}{*}{Camera} & \multirow{2}{*}{Resnet50} & \multirow{2}{*}{IPM} & \multirow{2}{*}{ART}  & \originalsplitname      & \moldlm & \moldre & \moldpc & \moldmap \\  
& & & & & & \oursplitname                                                                                                                          & \noldlm & \noldre & \noldpc & \noldmap \\ \cmidrule{7-\myendcolum} 
& \multirow{2}{*}{\begin{tabular}{c}MapTR \cite{liao2023maptr}\\ 2D\end{tabular}} & \multirow{2}{*}{Camera} & \multirow{2}{*}{Resnet50} & \multirow{2}{*}{GKT} & \multirow{2}{*}{DETR}  
&           \originalsplitname & {$64.0$}&{$62.8$} &  {$63.2$}&{$61.0$} & {$63.7$}&{$62.4$} & {$63.6$}&{$62.1$}  \\ 
& & & & & & \oursplitname      & {$50.0$}&{$45.2$} &  {$47.5$}&{$48.3$} & {$46.6$}&{$50.9$} & {$48.0$}&{$48.2$} \\ \cmidrule{7-\myendcolum} 
& \multirow{2}{*}{\begin{tabular}{c}MapTRv2 \cite{liao2023maptrv2}\\ 2D\end{tabular}}  & \multirow{2}{*}{Camera} & \multirow{2}{*}{Resnet50} & \multirow{2}{*}{LSS} & \multirow{2}{*}{DETR} & \originalsplitname           & \ooldlm & \ooldre & \ooldpc & \ooldmap \\  
& & & & & & \oursplitname                                                                                                                          & \poldlm & \poldre & \poldpc & \poldmap \\ \cmidrule{7-\myendcolum} 
& \multirow{2}{*}{\begin{tabular}{c}MapTRv2 \cite{liao2023maptrv2}\\ 3D\end{tabular}} & \multirow{2}{*}{Camera} & \multirow{2}{*}{Resnet50} & \multirow{2}{*}{LSS} & \multirow{2}{*}{DETR} & \originalsplitname           & \qoldlm & \qoldre & \qoldpc & \qoldmap \\  
& & & & & & \oursplitname                                                                                                                          & \roldlm & \roldre & \roldpc & \roldmap \\  \cmidrule{7-\myendcolum} 
& \multirow{2}{*}{\begin{tabular}{c}StreamMapNet \cite{yuan2023streammapnet}\\ 2D\end{tabular}} & \multirow{2}{*}{Camera} & \multirow{2}{*}{Resnet50} & \multirow{2}{*}{BEVFormer} & \multirow{2}{*}{DETR} & \originalsplitname       &    $58.3$ & $56.6$ & $63.9$ & $62.9$ & $62.7$ & $63.1$ & $61.7$ & $60.8$  \\  
& & & & & & \oursplitname                                                                                                                          & $52.7$ & $47.9$ & $50.0$ & $54.8$ & $49.4$ & $55.2$ & $50.7$ &  $52.6$  \\ 
\bottomrule 
\end{tabular}}
\caption{mAP comparison for methods trained on Original (\originalsplitname) and Near Extrapolation (\oursplitname) splits. All methods show a significant performance drop when trained and evaluated on Near. Autoregressive Transformer [ART], Object Detection with Transformer [DETR].}
\vspace{-15pt}
\label{tab:vec-split-performance}
\end{table*}

To demonstrate the geographical data leakage problem with the original splits of nuScenes and Argoverse 2, we evaluate the performance of state-of-the-art online mapping methods on both the original and proposed geographically disjoint data splits. Additionally, we re-validate studies performed in previous works.

Unless specified, no modifications have been made to the configurations of the evaluated methods, and we direct readers to the respective papers for specific training details.  Additionally, the performance is measured using standard practice in the respective field, \ie, Intersection over Union (IoU) for segmentation-based methods and mean average precision (mAP) \cite{li2022hdmapnet} for vector-based methods. For the latter, the average precision  $\text{AP}_\tau$ is calculated through thresholding the Chamfer distance between matched prediction/ground truth-pairs for the thresholds $\tau \in T, T = \{0.5, 1.0, 1.5\}$ to get
\begin{equation}
    \text{mAP} = \frac{1}{|T|}\sum_{\tau \in T}\text{AP}_\tau.
\end{equation}
We report mAP for individual object classes and their mean. 
\subsection{Data Leakage Effects}
\label{subsec:data-leakage-effects}
To investigate the effects of data leakage across data partitions we train several vector- and segmentation-based methods on both the Original and the geographically disjoint Near Extrapolation splits. The results for vector-based methods are reported in \cref{tab:vec-split-performance}. 
All evaluated methods see a significant performance drop when using geographically disjoint splits compared to the Original splits. 
The average performance decrement is more than $35$ mAP and $12$ mAP for nuScenes and Argoverse 2, respectively. 
Moreover, the effect is consistent over all lifting methods, sensor modalities, and decoders, but the ranking among the evaluated methods remains. 
The performance drop also remains consistent when adding lidar or considering the 3D geometry of the online map. 

The best-performing method (MapTRv2) on nuScenes using images as input drops from an mAP of $72.7$ to just $26.2$, showcasing a difference of $46.5$ mAP, when trained and evaluated appropriately. 
The drop is less pronounced, although still significant, on Argoverse 2, decreasing from $65.3$ to $55.2$ mAP. 
In general, the impact of the split is particularly distinct for methods trained on nuScenes. 
In light of these findings, we conclude that the smaller size of nuScenes, although convenient, makes it inadequate for training current online mapping methods. Moreover, although Argoverse 2 has more samples, it is somewhat surprising that algorithms trained on it still exhibit substantially improved generalization ability considering that the \emph{intra-set} overlap is also larger. 
We hypothesize that, despite the \emph{intra-set} overlap being greater, this overlap does not hinder training; instead, it possibly functions as natural and beneficial data augmentation similarly as synthetic augmentations have shown to be highly useful for image classification and object detection tasks \cite{Cubuk_2020_CVPR_Workshops}.

\begin{figure*}[ht]
  \centering
   \includegraphics[width=\linewidth, trim={0mm, 110mm, 0mm, 110mm}, clip]{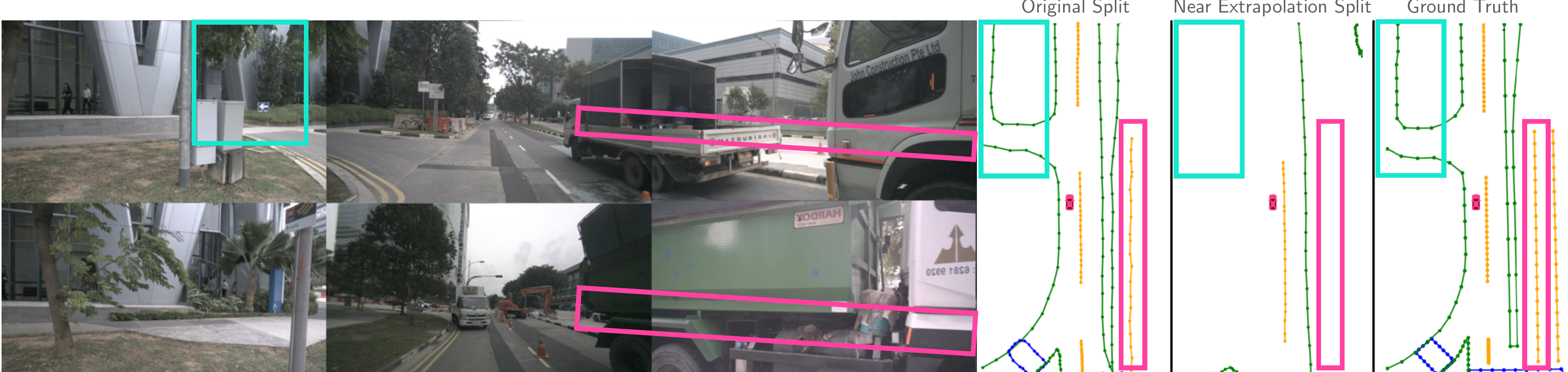}
   \caption{Predictions from MapTR \cite{liao2023maptr} trained on Original and Geographical splits along with the ground truth. Yellow lines denote (lane) dividers, green (road) boundaries, and blue pedestrian crossings. Note that, when trained on the Original split, the branch to the parallel road on the left (teal box) is not visible in any image, yet appears in the predicted map. Also, the divider on the opposing road to the right (pink box) is predicted very well. When training on geographically split data (here Near Extrapolation), this method fails to predict these.}
   \vspace{-0.45cm}
   \label{fig:mapt-qual-example}
\end{figure*}

To illustrate the significance of these numbers, a qualitative example is provided in \cref{fig:mapt-qual-example}, comparing MapTR \cite{liao2023maptr} predictions on a validation sample when trained on the Original and the Near Extrapolation nuScenes splits. 
Studying the figure carefully, we can see that the model trained on the original split accurately predicts the road edge (green line) even for areas completely occluded in the images, highlighted by the teal box. 
The method also predicts the lane divider (yellow line) on the opposing road through the trucks and barriers, as highlighted in the pink box. 
Given their single-shot nature and lack of consideration for previous sensor data, it is unreasonable that they can accurately predict road structures that are not visible or clearly indicated by other structures in the current view.
The method appears to learn to localize validation and test samples within the provided training map. 
This is not unique to this particular method or example, but rather a consequence of the overlap between train and evaluation sets. 
More qualitative examples are provided in \cref{sec:qualitative-results}.

For segmentation-based methods, \cref{tab:map-seg-multi} shows reduced performance when evaluated on geographically disjoint data. 
This suggests the impact extends beyond vectorized methods. 
Here, HDMapNet is kept the same as from the original paper and the other lifting techniques, namely Inverse Perspective Mapping (IPM), Cross-View Transformer (CVT) and Geometry-Guided Kernel Transformer (GKT) have been altered to predict the classes we are interested in.
Architectural details such as the image feature extraction backbone, Efficientnet-b4 \cite{tan2020efficientnet}, and segmentation decoder from SimpleBEV \cite{harley2022simplebev} are kept the same for all lifters for fair comparison. As for the vector-based online mapping methods, the drop is larger on nuScenes than Argoverse 2 and the ranking among methods remains largely unchanged. 

\begin{table}
    \small
    \centering
    \renewcommand{\arraystretch}{0.9}
    \setlength{\tabcolsep}{2pt} 
    \scalebox{0.8}{%
    \begin{tabular}{l c c cc cc cc cc cc}
    \hline 
    \toprule 
    & \multirow{2}{*}{Model} & \multirow{2}{*}{Split}  &\multicolumn{2}{c}{\lanemarkers} & \multicolumn{2}{c}{\roadedges} & \multicolumn{2}{c}{\zebras}& \multicolumn{2}{c}{Mean} \\ \cmidrule(lr){4-5} \cmidrule(lr){6-7} \cmidrule(lr){8-9} \cmidrule(lr){10-11}  
    & & & Val & Test & Val & Test & Val & Test & Val & Test \\ \midrule
    \multirow{4}{*}{\parbox[t]{2mm}{\multirow{4}{*}{\rotatebox[origin=c]{90}{nuScenes}}}} 
    &\multirow{2}{*}{GKT} & \originalsplitname.     & $25.8$ & $25.4$ & $25.6$ & $22.7$ & $6.2$  & $5.1$  & $19.2$ & $17.7$  \\ 
    && \oursplitname.      	                        & $12.5$ & $17.9$ & $12.6$ & $16.9$ & $1.4$  & $1.9$  & $8.8$  & $12.3$    \\ \cmidrule{3-11} 
    & \multirow{2}{*}{CVT} & \originalsplitname.    & $30.9$ & $30.1$ & $30.5$ & $25.5$ & $11.7$ & $7.6$  & $24.4$ & $21.1$  \\ 
    && \oursplitname.                               & $16.9$ & $11.6$ & $17.0$ & $10.3$ & $4.5$  & $1.1$  & $12.8$ & $7.7$   \\ \cmidrule{3-11} 
    & \multirow{2}{*}{IPM} & \originalsplitname.    & $46.8$ & $52.4$ & $50.0$ & $52.8$ & $26.8$ & $27.4$ & $41.2$ & $44.2$ \\ 
    && \oursplitname.                               & $29.6$ & $29.4$ & $36.2$ & $33.6$ & $16.2$ & $9.7$  & $27.4$ & $24.2$ \\ \cmidrule{3-11}
    &\multirow{2}{*}{HDMapNet} & \originalsplitname & $38.3$ & $47.5$ & $35.5$ & $41.2$ & $20.1$ & $27.6$ & $31.3$ & $38.8$ \\ 
    && \oursplitname                                & $8.6$  & $24.0$ & $22.1$ & $25.4$ & $10.6$ & $14.1$ & $17.1$ & $21.2$ \\ \midrule
    \multirow{4}{*}{\parbox[t]{2mm}{\multirow{4}{*}{\rotatebox[origin=c]{90}{Argoverse 2}}}} 
    &\multirow{2}{*}{GKT} & \originalsplitname.     & $37.3$ & $28.2$ & $31.4$ & $28.3$ & $10.3$ & $5.7$ & $26.3$ & $20.7$  \\ 
    && \oursplitname.      	                        & $32.7$ & $29.0$ & $26.2$ & $23.2$ & $8.7$  & $1.5$ & $22.5$ & $17.9$  \\ \cmidrule{3-11} 
    & \multirow{2}{*}{CVT} & \originalsplitname.    & $40.1$ & $29.4$ & $32.0$ & $29.3$ & $12.1$ & $7.1$  & $28.4$ & $21.9$ \\ 
    && \oursplitname.                               & $35.1$ & $31.7$ & $26.5$ & $28.5$ & $10.9$ & $1.4$ & $24.2$ & $20.5$  \\ \cmidrule{3-11} 
    & \multirow{2}{*}{IPM} & \originalsplitname.    & $58.5$ & $45.7$ & $50.6$ & $50.1$ & $33.4$ & $32.4$ & $47.5$ & $42.7$ \\ 
    && \oursplitname.                               & $50.5$ & $39.1$ & $44.1$ & $45.2$ & $29.7$ & $28.8$ & $41.5$ & $37.7$ \\ 
    \bottomrule 
    \end{tabular}}
    \caption{Map segmentation on the datasets, evaluating performance with IoU. The performance drops for all methods using the \oursplitname. versus \originalsplitname.. As for the vector-based OME, the drop is larger on nuScenes than Argoverse 2.}
    \vspace{-15pt}
    \label{tab:map-seg-multi}
\end{table}

Overall, the Near Extrapolation split yields a more consistent performance between the validation and test sets compared to the Original split. Suggesting a balanced distribution across sets and facilitates drawing reliable conclusions about hyperparameter choices. Ensuring that insights gained from the validation set generalize well to the test set. 

\subsection{Far Extrapolation Cross-validation}
We perform cross-validation using multiple folds of city-wise data partitioning to evaluate the performance under increased distribution shifts. \cref{tab:citywise-vector} shows the vector-based methods' performance for both nuScenes and Argoverse 2 using the Far Extrapolation splits. The performance drops even further using this split, emphasizing the difficulty current methods experience with extrapolating outside the training distribution. 
\begin{table}
\centering
\renewcommand{\arraystretch}{0.6}
\setlength{\tabcolsep}{2pt} 
\scalebox{0.8}{%
\begin{tabular}{l cc c c c c c}
\hline 
\toprule 
\multirow{2}{*}{\parbox[t]{2mm}{\multirow{2}{*}{\rotatebox[origin=c]{90}{}}}} & Model & Split & \lanemarkers & \roadedges & \zebras & Mean & CV \\ \cmidrule(lr){1-8} 
\multirow{6}{*}{\parbox[t]{2mm}{\multirow{6}{*}{\rotatebox[origin=c]{90}{nuScenes}}}}& 
\multirow{2}{*}{VectorMapNet} & A   & $7.6$  & $8.4$  & $5.9$  & $7.3$  & \multirow{2}{*}{$8.7$} \\ 
& & B                               & $11.9$ & $12.2$ & $6.1$  & $10.1$ & \\ \cmidrule{3-8} 
& \multirow{2}{*}{MapTR} & A        & $12.9$ & $21.1$ & $11.1$ & $15.0$ & \multirow{2}{*}{$14.9$}\\ 
& & B                               & $14.1$ & $24.6$ & $5.9$  & $14.9$ & \\ \cmidrule{3-8} 
& \multirow{2}{*}{MapTRv2} & A      & $18.6$ & $27.9$ & $18.9$ & $21.8$ & \multirow{2}{*}{$21.4$}\\ 
& & B                               & $22.4$ & $27.0$ & $13.3$ & $20.9$ & \\ \cmidrule{3-8} 
& \multirow{2}{*}{StreamMapNet} & A & $16.4$ & $22.7$ & $18.7$ & $19.3$ & \multirow{2}{*}{$21.3$}\\ 
& & B                               & $21.7$ & $30.6$ & $17.4$ & $23.2$ & \\ \cmidrule{1-8} 
\multirow{9}{*}{\parbox[t]{2mm}{\multirow{9}{*}{\rotatebox[origin=c]{90}{Argoverse 2}}}}
& \multirow{3}{*}{VectorMapNet} & A & $16.5$ & $19.0$ & $16.5$ & $21.1$  & \multirow{3}{*}{$24.0$}   \\ 
& &                               B & $29.4$ & $26.0$	& $20.5$ & $25.3$ \\ 
& &                               C & $28.0$ & $27.1$	& $22.1$ & $25.7$ \\ \cmidrule{3-8} 
& \multirow{3}{*}{MapTR} & A        & $41.7$ & $37.3$ & $34.7$ & $37.9$  & \multirow{3}{*}{$41.8$}\\ 
& & B                               & $47.5$ & $45.9$ & $40.2$ & $44.5$ \\
& & C                               & $41.2$ & $44.6$ & $43.3$ & $43.0$ \\ \cmidrule{3-8} 
& \multirow{3}{*}{MapTRv2} & A      & $42.2$ & $41.9$ & $37.4$ & $40.5$ & \multirow{3}{*}{$45.5$}\\ 
& & B                               & $53.4$ & $50.9$ & $42.0$ & $48.8$ \\
& & C                               & $50.3$ & $48.6$ & $42.6$ & $47.2$ \\ \cmidrule{3-8} 
& \multirow{3}{*}{StreamMapNet} & A & $43.4$ & $43.3$ & $40.2$ & $42.3$ & \multirow{3}{*}{$46.6$}\\ 
& & B                               & $51.0$ & $52.5$ & $46.7$ & $50.1$ \\
& & C                               & $42.7$ & $50.1$ & $49.4$ & $47.4$ \\
\bottomrule
\end{tabular}}
\caption{Vector-based methods' mAP on the Far Extrapolation folds and their corresponding cross-validation mean (CV).}
\vspace{-0.7cm}
\label{tab:citywise-vector}
\end{table}

\subsection{Sample Density}
We investigate the effect of the training set's sample density for both datasets.
As discussed in \cref{subsec:online-mapping-datasets} only the key-frame samples are typically used when training methods on nuScenes. 
We leverage the fact that \emph{all} samples have the vehicle poses required to extract the ground truth from the HD map and are able to use $4$ times as many samples for training. 
It is also forthright to downsample Argoverse 2 to the desired sample density by selecting every fourth sample, effectively simulating the sample density of nuScenes. 

The training schedule is adjusted such that the total number of optimizer steps is similar between the sparsely and densely sampled data. \cref{tab:nusc-dense-sampling} reports the result for MapTRv2. For nuScenes there is a distinguished increase in performance on the original split, achieving up to a 24.9 mAP improvement on the validation set by utilizing the densely sampled data. 
However, the method's performance using the Near Extrapolation split sees only marginal improvement (max increase of 0.7 mAP). Argoverse 2 sees smaller differences between the sample densities.
\begin{table}
    \centering
    \renewcommand{\arraystretch}{0.8} 
    \setlength{\tabcolsep}{2pt} 
    \scalebox{0.80}{%
    \begin{tabular}{l c c cc cc cc cc}\hline 
    \toprule 
     & \multirow{2}{*}{Split} & \multirow{2}{*}{\begin{tabular}{c}Train\\ Sampling\end{tabular}} & \multicolumn{2}{c}{Divider} & \multicolumn{2}{c}{Boundary} & \multicolumn{2}{c}{Crossing} & \multicolumn{2}{c}{Mean} \\ \cmidrule(lr){4-5} \cmidrule(lr){6-7} \cmidrule(lr){8-9} \cmidrule(lr){10-11}  
     & & & Val & Test & Val & Test & Val & Test & Val & Test \\ \midrule
     \multirow{2}{*}{\parbox[t]{2mm}{\multirow{2}{*}{\rotatebox[origin=c]{90}{nuScenes}}}} 
     & \multirow{2}{*}{Orig.} & Sparse & $61.8$ & $77.3$ & $63.7$ & $70.9$ & $59.1$ & $69.8$ & $61.5$ & $72.7$ \\ 
                            & & Dense & $90.3$ & $89.7$ & $84.4$ & $84.5$ & $86.4$ & $88.3$ & $86.4$ & $87.5$\\ \cmidrule{3-11}
     & \multirow{2}{*}{\oursplitname}  & Sparse & $20.9$ & $23.4$ & $32.6$ & $40.5$ & $26.5$ & $14.8$ & $26.7$ & $26.2$ \\ 
                            & & Dense  & $21.7$ & $24.1$ & $34.1$ & $40.9$ & $25.7$ & $15.9$ & $27.2$ & $26.9$ \\ \midrule 
     \multirow{2}{*}{\parbox[t]{2mm}{\multirow{2}{*}{\rotatebox[origin=c]{90}{Argo 2}}}} 
     & \multirow{2}{*}{Orig.} 
                              & Sparse   & $67.7$ & $64.9$ & $63.6$ & $59.8$ & $58.8$ & $57.1$ & $63.4$ & $60.6$ \\ 
                            & & Dense & $71.7$ & $68.9$ & $67.0$ & $63.8$ & $64.5$ & $63.1$ & $67.7$ & $65.3$ \\ \cmidrule{3-11}
     & \multirow{2}{*}{\oursplitname}  
                              & Sparse   & $55.4$ & $54.0$ & $48.2$ & $52.5$ & $44.8$ & $51.5$ & $49.5$ & $52.6$ \\ 
                            & & Dense & $58.4$ & $56.6$ & $51.3$ & $53.5$ & $49.7$ & $55.6$ & $53.1$ & $55.2$ \\
     \bottomrule
    \end{tabular}}
    \caption{MapTRv2 mAP trained on varying dataset density. For nuScenes, dense sampling boosts performance by up to $24.9$ on the Orig., while Near only sees minor improvements.}
    \label{tab:nusc-dense-sampling}
    \vspace{-0.5cm}
\end{table}

\subsection{Re-validation}
As original works validate design choices and hyperparameters on poorly separated data, it is highly relevant to revisit these results to check applicability to proposed split.
Though not covering all prior tests, we highlight some interesting observations on the Near Extrapolation split.
Hyperparameter-search shows minor differences, see \cref{sec:extended-experiments}, while new conclusions emerge regarding lifting method and auxiliary tasks.

\vspace{-0.37cm}
\paragraph{Lifting methods}
\label{par:lifting-method}
In \cite{liao2023maptr}, an ablation study investigates the impact of various lifters for MapTR, with GKT yielding the best results. 
However, upon re-running this test (see \cref{tab:maptr-lifter-ablation}), we observe a contradiction to the previous findings. The BEVFormer lifter slightly outperforms GKT. Nonetheless, the differences between the lifters are marginal, making it challenging to determine the superiority of any specific lifter.

\vspace{-0.37cm}
\paragraph{Auxiliary tasks}
For MapTRv2 \cite{liao2023maptrv2}, we re-run the ablation studies on the proposed auxiliary tasks in \cref{tab:maptrv2-auxtask-ablation}. 
For nuScenes, we note that, in contrast to conclusions based on the original split, the addition of depth supervision does not yield a significant performance boost. 
Additionally, one can infer that it is only when all auxiliary tasks are combined that the improvement becomes apparent. 
However, the effects of the additional tasks are smaller than initially concluded when training on the original split. 
Considering Argoverse 2, there are bigger differences between the performance among the auxiliary tasks. 
Similarly to the re-validation on nuScenes, the effectiveness of, \eg, depth supervision is not as striking as previously advertised. 

\def\NoTrainMaptrGktOgLm{$51.0$}
\def\NoTrainMaptrGktOgRe{$52.6$}
\def\NoTrainMaptrGktOgPc{$43.0$}
\def\NoTrainMaptrGktOgMap{$48.8$}
\def\NoTrainMaptrGktNewLm{$16.0$}
\def\NoTrainMaptrGktNewRe{$26.7$}
\def\NoTrainMaptrGktNewLc{$14.4$}
\def\NoTrainMaptrGktNewMap{$19.0$}

\def\NoTrainMaptrBevformerOgLm{$49.7$}
\def\NoTrainMaptrBevformerOgRe{$53.5$}
\def\NoTrainMaptrBevformerOgPc{$40.2$}
\def\NoTrainMaptrBevformerOgMap{$47.8$}
\def\NoTrainMaptrBevformerNewLm{$16.2$}
\def\NoTrainMaptrBevformerNewRe{$28.2$}
\def\NoTrainMaptrBevformerNewLc{$17.5$}
\def\NoTrainMaptrBevformerNewMap{$20.6$}

\def\NoTrainMaptrBevpoolOgLm{$52.1$}
\def\NoTrainMaptrBevpoolOgRe{$52.4$}
\def\NoTrainMaptrBevpoolOgPc{$45.4$}
\def\NoTrainMaptrBevpoolOgMap{$50.0$}
\def\NoTrainMaptrBevpoolNewLm{$17.3$}
\def\NoTrainMaptrBevpoolNewRe{$27.7$}
\def\NoTrainMaptrBevpoolNewLc{$18.0$}
\def\NoTrainMaptrBevpoolNewMap{$21.0$}

\def\NoTrainArgoMaptrGktOgLm{$64.0$}
\def\NoTrainArgoMaptrGktOgRe{$63.2$}
\def\NoTrainArgoMaptrGktOgPc{$63.7$}
\def\NoTrainArgoMaptrGktOgMap{$63.6$}
\def\NoTrainArgoMaptrGktNewLm{$50.0$}
\def\NoTrainArgoMaptrGktNewRe{$47.5$}
\def\NoTrainArgoMaptrGktNewLc{$46.6$}
\def\NoTrainArgoMaptrGktNewMap{$48.0$}

\def\NoTrainArgoMaptrBevformerOgLm{$63.7$}
\def\NoTrainArgoMaptrBevformerOgRe{$63.8$}
\def\NoTrainArgoMaptrBevformerOgPc{$63.0$}
\def\NoTrainArgoMaptrBevformerOgMap{$63.5$}
\def\NoTrainArgoMaptrBevformerNewLm{$49.5$}
\def\NoTrainArgoMaptrBevformerNewRe{$47.2$}
\def\NoTrainArgoMaptrBevformerNewLc{$46.3$}
\def\NoTrainArgoMaptrBevformerNewMap{$47.7$}

\def\NoTrainArgoMaptrBevpoolOgLm{$64.8$}
\def\NoTrainArgoMaptrBevpoolOgRe{$65.2$}
\def\NoTrainArgoMaptrBevpoolOgPc{$61.9$}
\def\NoTrainArgoMaptrBevpoolOgMap{$64.0$}
\def\NoTrainArgoMaptrBevpoolNewLm{$49.7$}
\def\NoTrainArgoMaptrBevpoolNewRe{$47.3$}
\def\NoTrainArgoMaptrBevpoolNewLc{$45.1$}
\def\NoTrainArgoMaptrBevpoolNewMap{$47.3$}

\begin{table}
    \centering
    \renewcommand{\arraystretch}{0.8} 
    \setlength{\tabcolsep}{2pt} 
    \scalebox{0.80}{%
    \begin{tabular}{l cc c c c c}\hline 
    \toprule 
    \multirow{2}{*}{\parbox[t]{2mm}{\multirow{2}{*}{\rotatebox[origin=c]{90}{}}}} & Lifting & Split & \lanemarkers & \roadedges & \zebras & \multicolumn{1}{c}{Mean} \\ \cmidrule{1-7} 
    \multirow{4}{*}{\parbox[t]{2mm}{\multirow{4}{*}{\rotatebox[origin=c]{90}{nuScenes}}}} & 
    \multirow{2}{*}{GKT} & 
        \originalsplitname               & \NoTrainMaptrGktOgLm & \NoTrainMaptrGktOgRe & \NoTrainMaptrGktOgPc & \NoTrainMaptrGktOgMap \\ 
    & & \oursplitname & \NoTrainMaptrGktNewLm & \NoTrainMaptrGktNewRe & \NoTrainMaptrGktNewLc & \NoTrainMaptrGktNewMap \\ \cmidrule{3-7} 
    & \multirow{2}{*}{BEVFormer} &
        \originalsplitname               & \NoTrainMaptrBevformerOgLm & \NoTrainMaptrBevformerOgRe & \NoTrainMaptrBevformerOgPc & \NoTrainMaptrBevformerOgMap \\ 
    & & \oursplitname          & \NoTrainMaptrBevformerNewLm & \NoTrainMaptrBevformerNewRe & \NoTrainMaptrBevformerNewLc & \NoTrainMaptrBevformerNewMap \\ \cmidrule{3-7} 
    & \multirow{2}{*}{LSS} &
        \originalsplitname               & \NoTrainMaptrBevpoolOgLm & \NoTrainMaptrBevpoolOgRe & \NoTrainMaptrBevpoolOgPc & \NoTrainMaptrBevpoolOgMap \\ 
    & & \oursplitname          & \NoTrainMaptrBevpoolNewLm & \NoTrainMaptrBevpoolNewRe & \NoTrainMaptrBevpoolNewLc & \NoTrainMaptrBevpoolNewMap \\ \cmidrule{1-7} 
    \multirow{4}{*}{\parbox[t]{2mm}{\multirow{4}{*}{\rotatebox[origin=c]{90}{Argoverse 2}}}} & 
    \multirow{2}{*}{GKT} & 
        \originalsplitname               & \NoTrainArgoMaptrGktOgLm & \NoTrainArgoMaptrGktOgRe & \NoTrainArgoMaptrGktOgPc & \NoTrainArgoMaptrGktOgMap \\ 
    & & \oursplitname          & \NoTrainArgoMaptrGktNewLm & \NoTrainArgoMaptrGktNewRe & \NoTrainArgoMaptrGktNewLc & \NoTrainArgoMaptrGktNewMap \\ \cmidrule{3-7} 
    & \multirow{2}{*}{BEVFormer} &
        \originalsplitname               & \NoTrainArgoMaptrBevformerOgLm & \NoTrainArgoMaptrBevformerOgRe & \NoTrainArgoMaptrBevformerOgPc & \NoTrainArgoMaptrBevformerOgMap \\ 
    & & \oursplitname          & \NoTrainArgoMaptrBevformerNewLm & \NoTrainArgoMaptrBevformerNewRe & \NoTrainArgoMaptrBevformerNewLc & \NoTrainArgoMaptrBevformerNewMap \\ \cmidrule{3-7} 
    & \multirow{2}{*}{LSS} &
        \originalsplitname               & \NoTrainArgoMaptrBevpoolOgLm & \NoTrainArgoMaptrBevpoolOgRe & \NoTrainArgoMaptrBevpoolOgPc & \NoTrainArgoMaptrBevpoolOgMap \\ 
    & & \oursplitname          & \NoTrainArgoMaptrBevpoolNewLm & \NoTrainArgoMaptrBevpoolNewRe & \NoTrainArgoMaptrBevpoolNewLc & \NoTrainArgoMaptrBevpoolNewMap \\ 
    \bottomrule
    \end{tabular}}
    \caption{Validation mAP for lifting methods in MapTR. Marginal differences between the lifters make it challenging to establish the superiority of any particular method.}
    \label{tab:maptr-lifter-ablation}
\end{table}
\def\NuscNoAux{$25.3$}
\def\NuscBEV{$25.8$}
\def\NuscDepthPV{$26.1 $}
\def\NuscDepthBEV{$25.9 $}
\def\NuscPVBEV{$25.5 $}
\def\NuscAll{$\mathbf{26.7}$}
\def\NuscDepth{$25.9$}

\def\ArgoNoAux{$48.8$}
\def\ArgoBEV{$50.2$}
\def\ArgoDepthPV{$50.6$}
\def\ArgoDepthBEV{$50.9$}
\def\ArgoPVBEV{$51.5$}
\def\ArgoAll{$\mathbf{53.1}$}
\def\ArgoDepth{$48.8$}
\begin{table}
    \centering
    \renewcommand{\arraystretch}{0.8} 
    \setlength{\tabcolsep}{2pt} 
    \scalebox{0.80}{%
    \begin{tabular}{ccc cc c}\hline 
    \toprule 
    \multirow{2}{*}{Depth} & \multirow{2}{*}{Seg$^{\text{PV}}$} & \multirow{2}{*}{Seg$^{\text{BEV}}$} & \multicolumn{2}{c}{nuScenes} & Argoverse 2 \\ \cmidrule(lr){4-5} \cmidrule(lr){6-6}
     & & & Orig. & Geo. & Geo.\\ \midrule
     & &   &                    $56.6$ & \NuscNoAux &\ArgoNoAux \\
     \cmark &  &  &             $59.8$ &\NuscDepth & \ArgoDepth \\ 
     \cmark & \cmark & &        $60.5$ &\NuscDepthPV & \ArgoDepthPV\\ 
     \cmark & & \cmark&         $61.0$ & \NuscDepthBEV &  \ArgoDepthBEV \\ 
     & \cmark & \cmark&         $59.2$ &\NuscPVBEV & \ArgoPVBEV \\ 
     \cmark & \cmark & \cmark & $\mathbf{61.5}$ & \NuscAll & \ArgoAll \\ 
     \bottomrule
    \end{tabular}}
    \caption{Validation mAP for auxiliary tasks in MapTRv2. nuScenes Orig. numbers are from \cite{liao2023maptrv2}. \oursplitname yields a smaller performance boost of auxiliary tasks. For Argoverse 2, the differences are greater, but inconsistent with the result on Orig..}
    \vspace{-0.5cm}
    \label{tab:maptrv2-auxtask-ablation}
\end{table}

\section{Conclusion}
\label{sec:conclusion}
We propose and employ geographically disjoint splits of the most used datasets, revealing that the performance of state-of-the-art online mapping methods is significantly lower than previously reported. 
We argue that these splits offer a more accurate measure of how well online mapping methods generalize to new geographic areas. While the Near Extrapolation split acts as a drop-in replacement to the original splits, we urge the community to target the Far Extrapolation setting moving forward.

Even though performance numbers have decreased for all methods with these splits, the ranking between methods remains largely the same. 
The performance disparity is more pronounced on nuScenes than Argoverse 2. 
However, our follow-up re-validation experiments have revealed new insights, diverging from conclusions based on the original split. 
Notably, the impact of the lifting method and the support from auxiliary tasks, \eg depth supervision, on performance appears less substantial or follows a different trajectory than initially perceived.

In summary, online mapping remains a formidable challenge, and to make substantial progress, we must anchor our conclusions in fair evaluations based on clean data splits. 
We look forward to what innovations will come from the improved evaluation ability with the release of our geographically disjoint data splits.

\vspace{-0.2cm}
\paragraph{Acknowledgements:}
This work was partially supported by the Wallenberg AI, Autonomous Systems and Software Program (WASP) funded by the Knut and Alice Wallenberg Foundation. Computational resources were provided by the National Academic Infrastructure for Supercomputing in Sweden (NAISS) at \href{https://www.nsc.liu.se/}{NSC Berzelius} and \href{https://www.c3se.chalmers.se/about/Alvis/}{C3SE Alvis} partially funded by the Swedish Research Council through grant agreement no. 2022-06725.
\vspace{-0.75cm}
{
    \small
    \bibliographystyle{ieeenat_fullname}
    \bibliography{main}
}

\clearpage
\setcounter{page}{1}
\maketitlesupplementary
\appendix

\section{Partially Overlapping Maps}
\label{sec:overlapping-maps}
\begin{figure}
  \centering
   \includegraphics[width=0.6\linewidth, trim={0mm, 5mm, 340mm, 40mm}, clip]{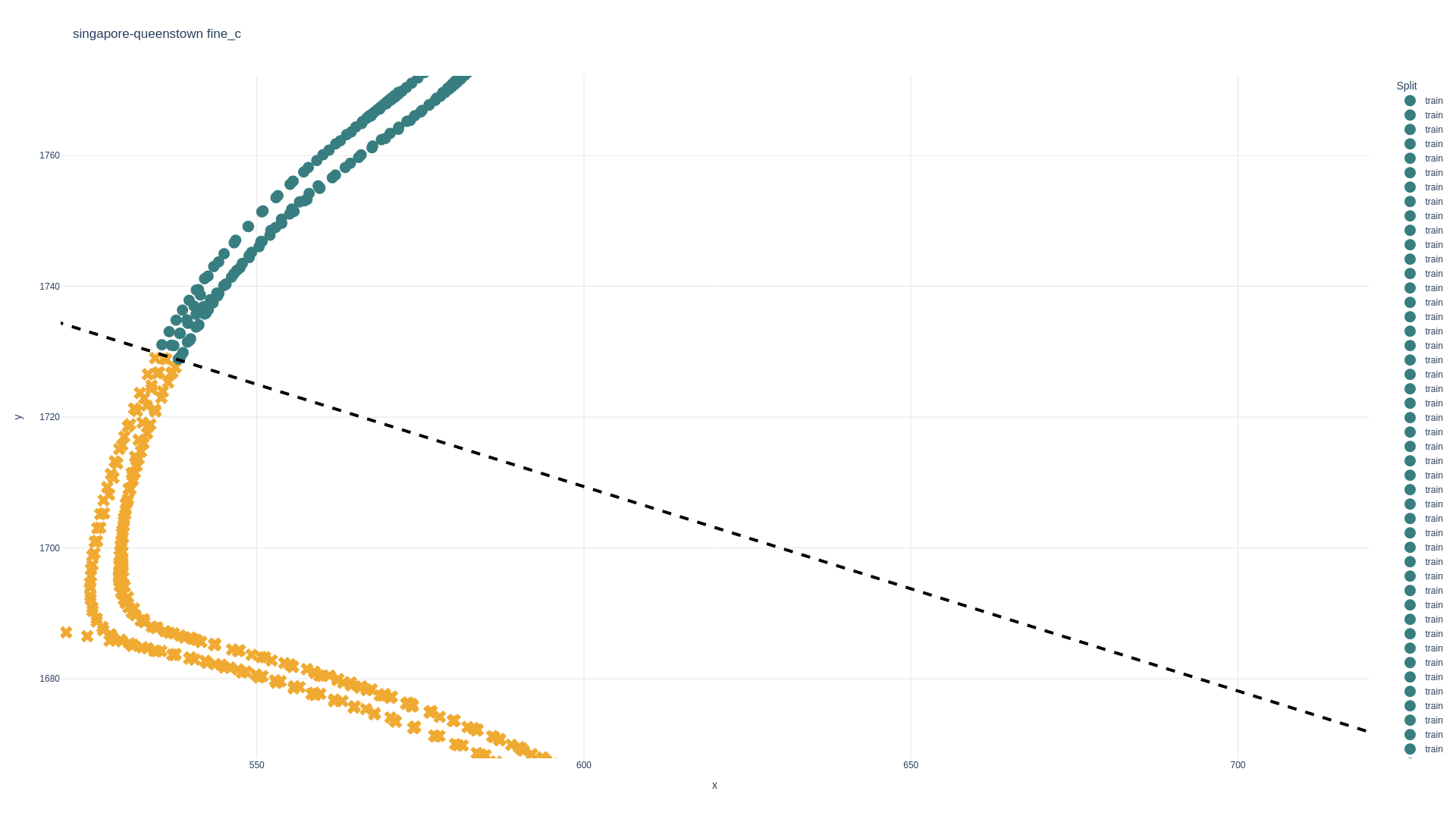}
   \caption{In this example from Singapore Queenstown, the individual samples from a few sequences are divided into training (green) and validation (orange) according to the cut-off border. Some samples from a sequence are put in the training set, whereas the remaining are put in the validation set. The samples close to the dotted black cut-off line are the remaining possible data-leakage samples when using our proposed Near Extrapolation split. }
   \label{fig:nusc-split-scenes}
\end{figure}

Splitting nuScenes for Near Extrapolation on a sequence level requires grouping large areas with similar zone classes together and putting them in a single set, as seen in \cite{philion2020lift}. This is due to the entangled nature of the sequences where many partially overlap. Instead, we assign each sample individually to a set when a sequence straddles the boundary between two sets (\eg train and val in \cref{fig:nusc-split-scenes}). We divide the sequence at the boundary, creating two separate partial sequences each with preserved temporal consistency. 
This maintains the usefulness for object detection and keeps the possibility of using the data for temporal fusion, where having consecutive samples is important. We have kept the number of sequences being cut into multiple parts as low as possible, making the cuts, when necessary, across the road's driving direction. 

The sequences in the Argoverse 2 dataset are more spread out compared to nuScenes, and a balanced sequence-wise split is possible to obtain. There is thus no impact on usability for object detection, object tracking, and other temporal fusion applications for the Argoverse 2 split.

\begin{figure}
    \centering
    \includegraphics[width=\linewidth]{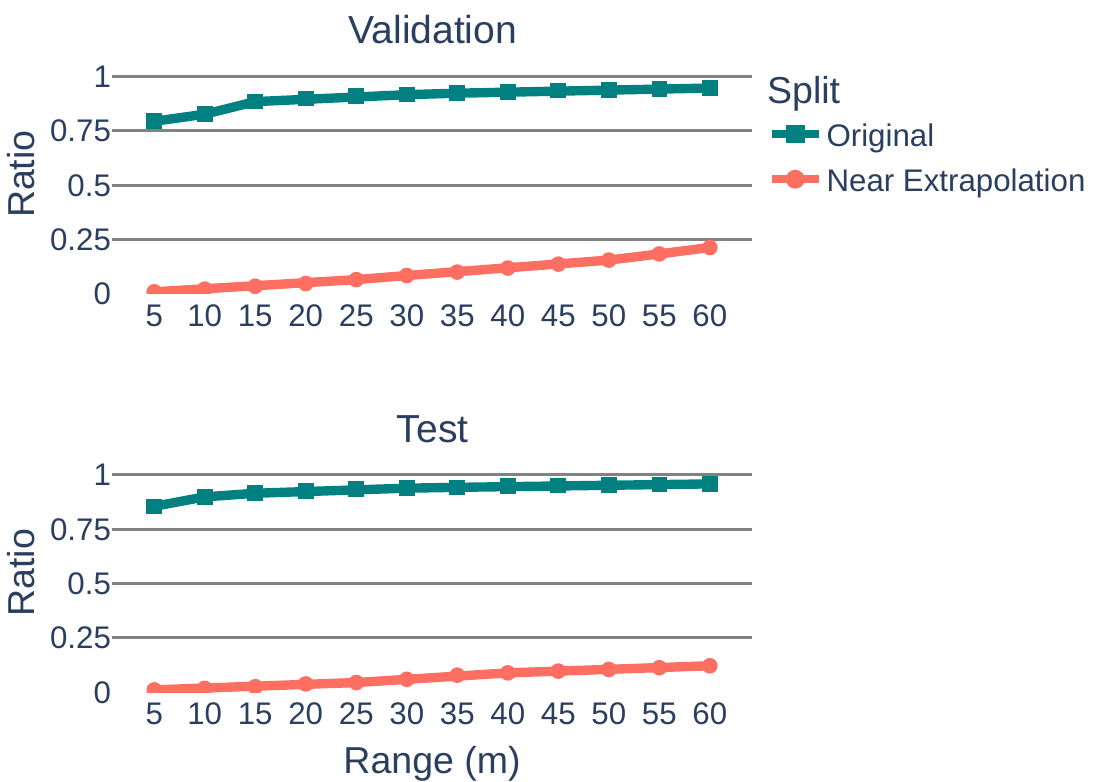}
    \caption{Ratios of validation and test samples within a certain range of training samples for nuScenes. The Geographically disjoint Near Extrapolation split has negligible overlap compared to the, greatly overlapping, Original split.}
    \label{fig:nusc-overlap-bar}
\end{figure}

\begin{figure}
    \centering
    \includegraphics[width=\linewidth]{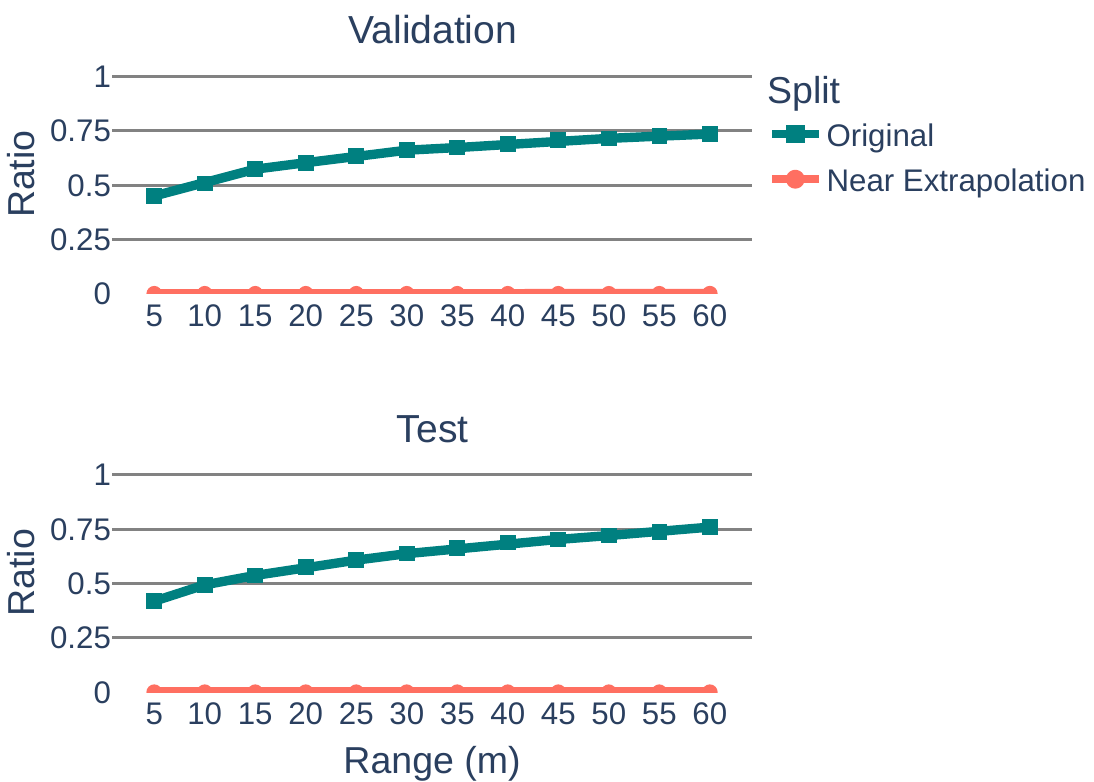}
    \caption{Ratios of validation and test samples within a certain range of training samples for Argoverse 2. The Geographically disjoint Near Extrapolation split has no overlap compared to the, greatly overlapping, Original split.}
    \label{fig:argo-overlap-bar}
\end{figure}

Splitting the data geographically ensures that there is no overlap in poses between the different sets. However, as online mapping methods typically predict $30$m in front and to the rear there will still be some overlap in the ground truth maps among the samples close to the cut-off border. To see the effects of the remaining overlap in the geographical split of nuScenes we run experiments where the validation and test samples closer than $60$ m to a training sample have been filtered out. \cref{tab:filter-out-close} demonstrates that these samples have a negligible impact on performance. Furthermore, \cref{fig:nusc-overlap-bar} shows how the ratio of validation and test samples that are close to a training sample changes with range. For completeness \cref{fig:argo-overlap-bar} displays the same information on Argoverse 2.
\begin{table}
\small
\centering
\scalebox{0.8}{%
\begin{tabular}{l cc cc cc cc}\hline 
\toprule 
\multirow{2}{*}{Split} & \multicolumn{2}{c}{HDMapNet} & \multicolumn{2}{c}{VectorMapNet} & \multicolumn{2}{c}{MapTR} & \multicolumn{2}{c}{MapTRv2} \\ \cmidrule(lr){2-3} \cmidrule(lr){4-5} \cmidrule(lr){6-7} \cmidrule(lr){8-9}
 & Val & Test & Val & Test & Val & Test & Val & Test \\ \midrule 
\oursplitname & $17.1$ & $21.2$ & $14.0$ & $18.2$ & $19.0$ & $19.7$ & $26.7$ & $26.2$\\ 
\oursplitname$\notin$ & $17.0$ & $21.4$ & $14.5$ & $18.3$ & $19.0$ & $19.6$ & $26.5$ & $25.8$\\ 
\bottomrule 
\end{tabular}}
\caption{Evaluating the predictions from validation and test sets in the nuScenes' Near Extrapolation split, where samples closer than 60m to a training sample have been removed (indicated by $\notin$). It can be seen that the impact on performance is negligible. Metrics are IoU for HDMapNet, and mAP for MapTRv2 and VectorMapNet. }
\label{tab:filter-out-close}
\end{table}


\section{Additional Data Attributes}
\label{sec:additional-data-attributes}
In this section, we further display the splitting, the number of samples in discretized maps, and different zone classes (\eg residential, commercial, and industrial). 

\begin{figure}[t]
    \centering
    \includegraphics[width=\linewidth, trim={1mm, 0, 1mm, 0}, clip]{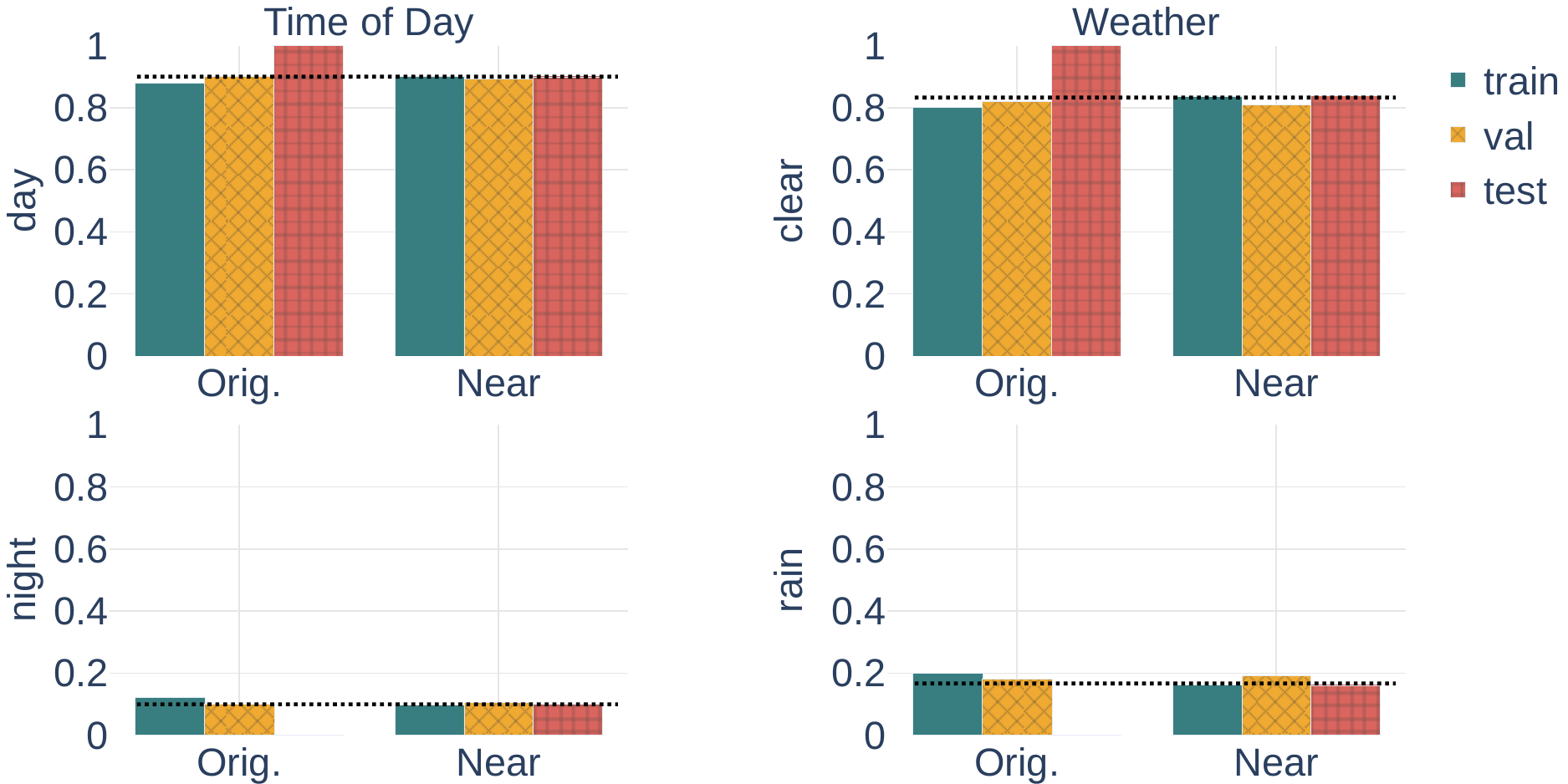}
    \caption{Ratios of weather conditions (clear and rain) as well as time of day (day and night) on the nuScenes dataset. The black dashed lines are the respective ratios over the full dataset.}
    \label{fig:nusc_night_rain}
\end{figure}

\begin{figure*}[h]
  \centering
   \includegraphics[width=\linewidth, trim={40mm, 0mm, 40mm, 0mm}, clip]{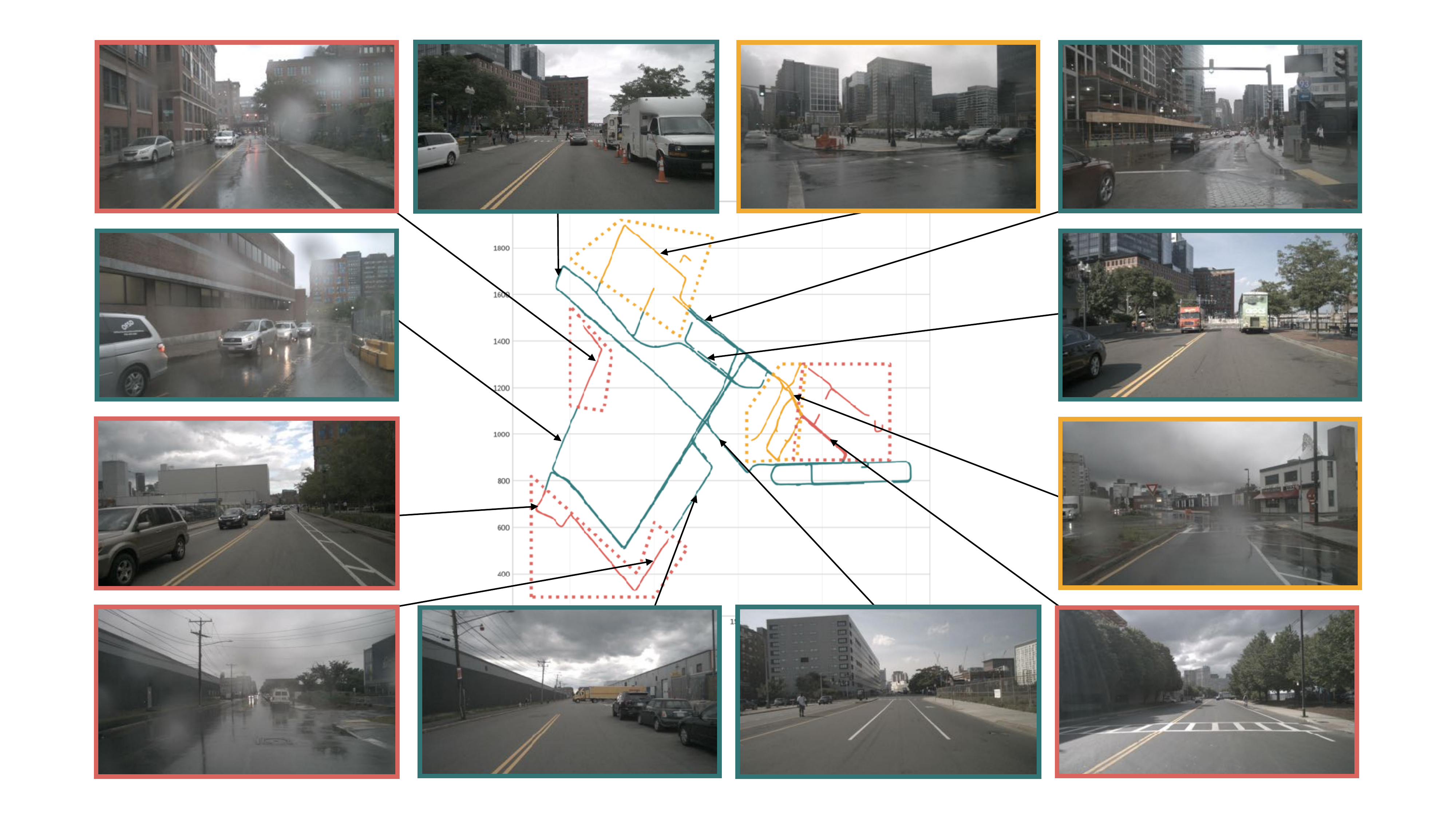}
   \caption{Selected poses from the Boston Seaport map in nuScenes dataset, with marked training (green), validation (blue), and test (red) poses according to the Near Extrapolation split. Dotted polygons mark the boundaries of the validation and test zones. To ensure diversity in zone types within each set, regions from various parts of the city are included. The industrial zones in the south and south-eastern areas have different attributes than the commercial and residential zones in the north-western part.}
   \label{fig:nusc-boston-thumbnails}
\end{figure*}

\begin{figure*}[h]
    \centering
    \begin{subfigure}[b]{0.45\textwidth}
    \centering
    \includegraphics[width=\linewidth]{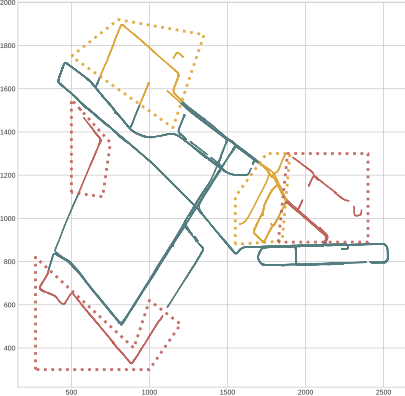}
    \caption{Boston Seaport}
    \label{fig:nusc-boston-seaport}
    \end{subfigure}
    \begin{subfigure}[b]{0.45\textwidth}
    \centering
    \includegraphics[width=\linewidth]{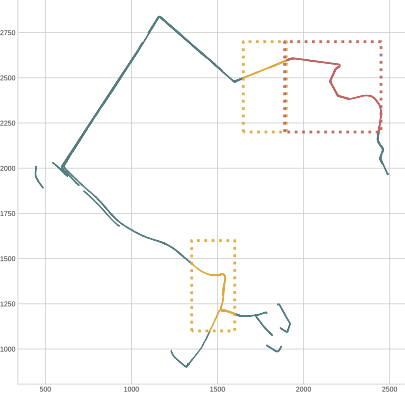}
    \caption{Singapore Hollandvillage}
    \label{fig:nusc-singapore-hollandvillage}
    \end{subfigure}
    
    \begin{subfigure}[b]{0.45\textwidth}
    \centering
    \includegraphics[width=\linewidth]{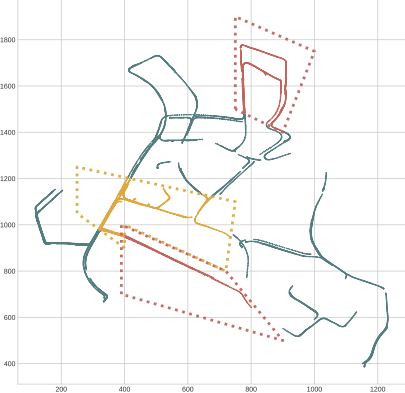}
    \caption{Singapore Onenorth}
    \label{fig:nusc-singapore-onenorth}
    \end{subfigure}
    \begin{subfigure}[b]{0.45\textwidth}
    \centering
    \includegraphics[width=\linewidth]{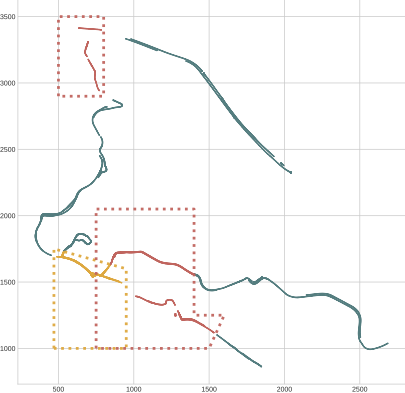}
    \caption{Singapore Queenstown}
    \label{fig:nusc-singapore-queenstown}
    \end{subfigure}
\caption{Positions of samples in the nuScenes dataset, with the geographical areas of the Near Extrapolation split outlined by dotted polygons. Training, validation, and test sets are distinguished by green, orange, and red colors, respectively. Areas from various parts of the cities are present in each set.}
\label{fig:nusc-sample-pos}
\end{figure*}

\begin{figure*}[h]
    \centering
    \begin{subfigure}[b]{0.45\textwidth}
        \centering
        \includegraphics[width=\linewidth, trim={0mm, 0mm, 0mm, 18mm}, clip]{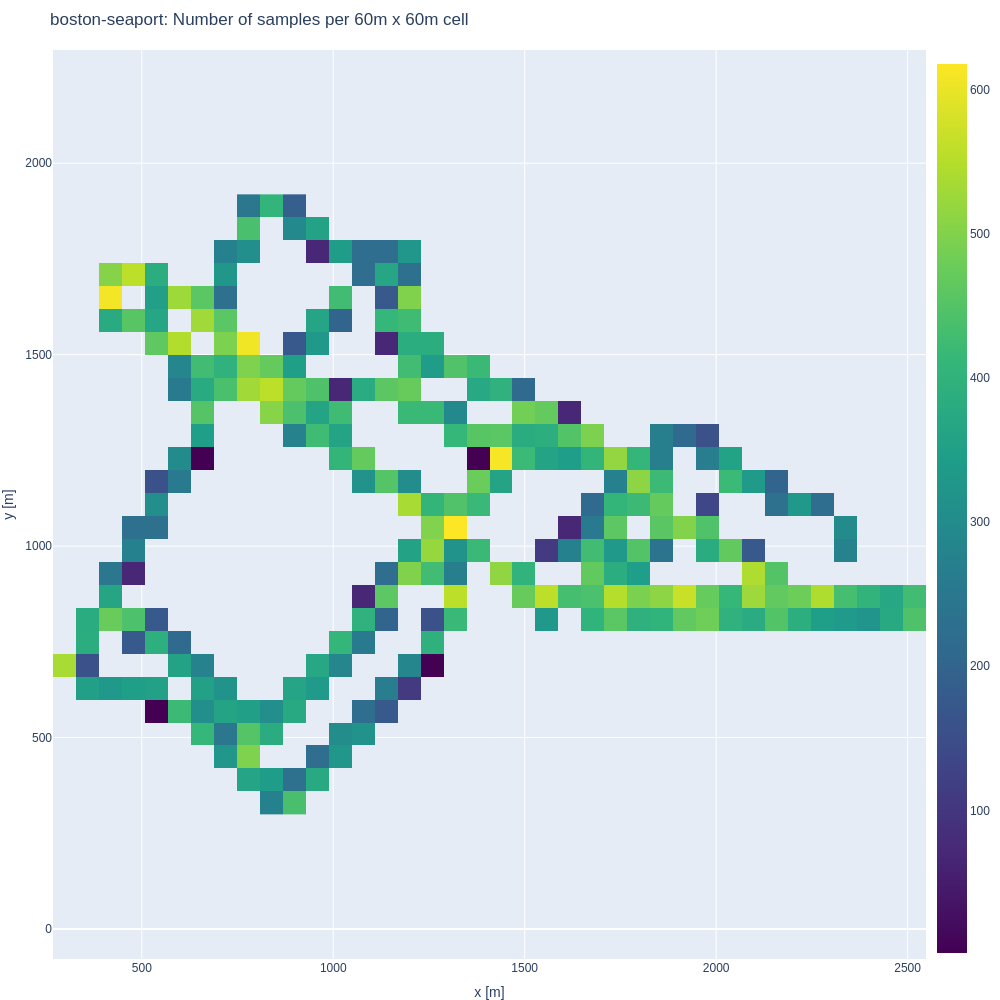}
        \caption{Boston Seaport}
    \end{subfigure}
    \begin{subfigure}[b]{0.45\textwidth}
    \centering
        \includegraphics[width=\linewidth, trim={0mm, 0mm, 0mm, 18mm}, clip]{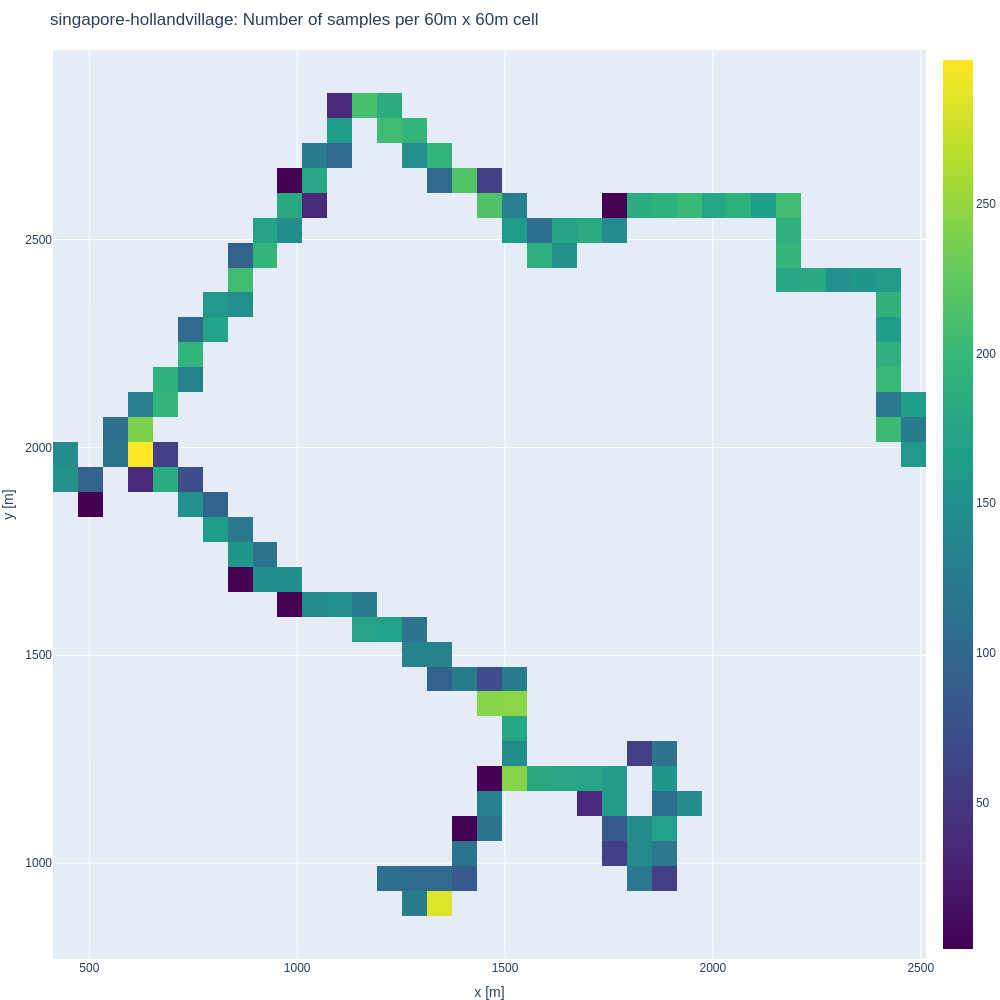}
        \caption{Singapore Hollandvillage}
    \end{subfigure}
    
    \begin{subfigure}[b]{0.45\textwidth}
        \centering
        \includegraphics[width=\linewidth, trim={0mm, 0mm, 0mm, 18mm}, clip]{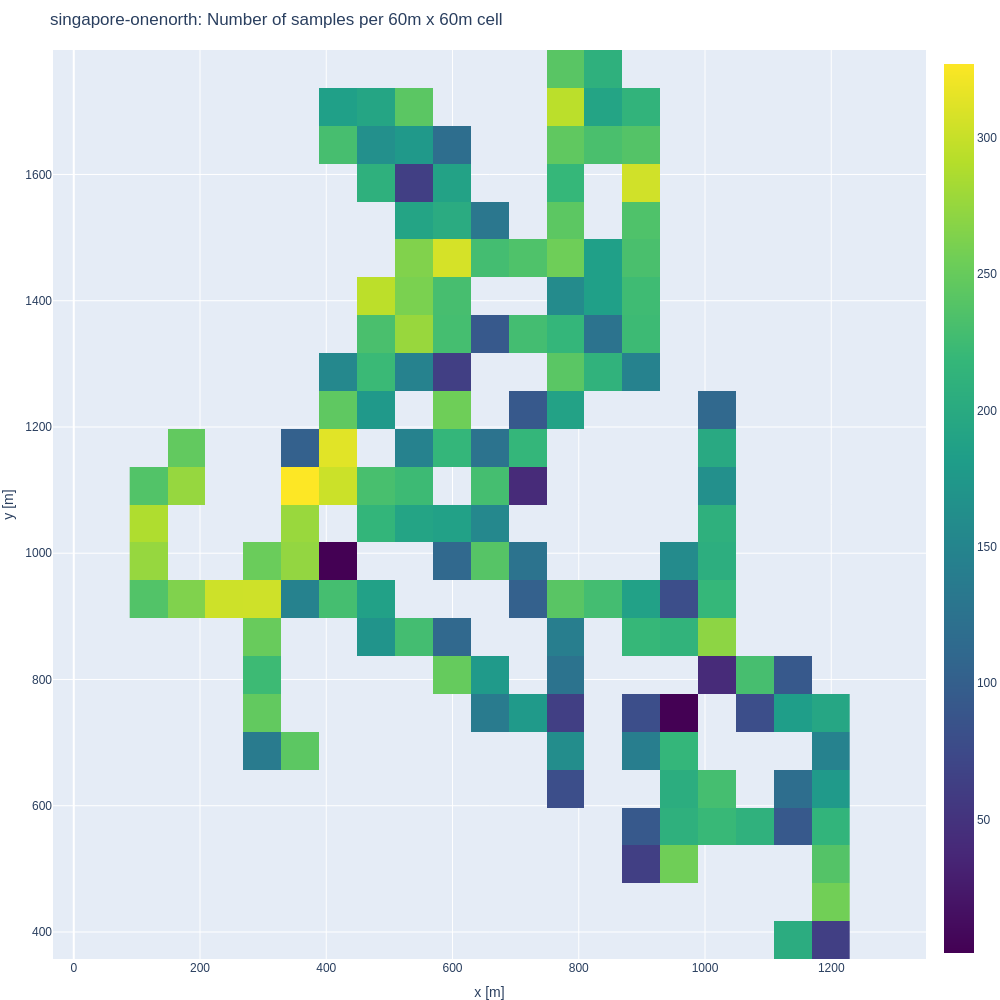}
        \caption{Singapore Onenorth}
    \end{subfigure}
    \begin{subfigure}[b]{0.45\textwidth}
        \centering
        \includegraphics[width=\linewidth, trim={0mm, 0mm, 0mm, 18mm}, clip]{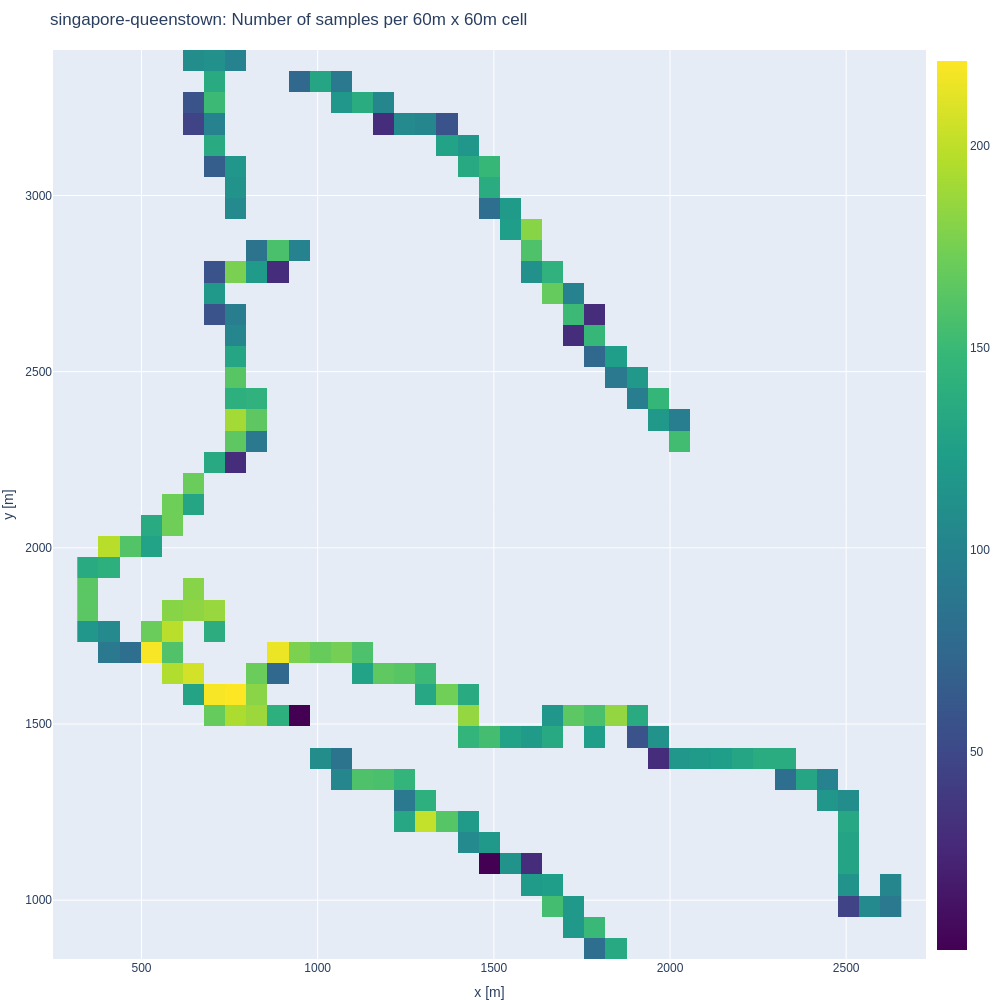}
        \caption{Singapore Queenstown}
    \end{subfigure}
   \caption{Heatmaps depicting the distribution of samples within 60m cells in the nuScenes dataset, revealing a high amount of samples in many cells, especially concentrated within crossings.}
   \label{fig:nusc-heatmaps}
\end{figure*}

\paragraph{nuScenes}
\cref{fig:nusc_night_rain} details that the Near Extrapolation split is balanced across all attributes. This allows for conducting experiments and drawing conclusions on a well-defined dataset. Further, \cref{fig:nusc-boston-thumbnails} depicts example images and their position on the map for Boston Seaport. The industrial zones in the south and south-eastern areas have different attributes, \eg type of buildings, lane widths, number of lanes, and frequency of pedestrian crossings, than the commercial and residential zones in the north-western part. It is thus important that these zones are represented in all sets for a fair evaluation of trained methods.
\cref{fig:nusc-sample-pos} showcases the regions where samples are allocated in each set for all cities. Each set incorporates regions from different parts of the cities to promote diversity. The heatmaps in \cref{fig:nusc-heatmaps} depict the distribution of samples within each 60m cell. One can, for instance, observe a concentration of samples in crossings. 

\begin{figure}[t]
    \centering
    \includegraphics[width=\linewidth]{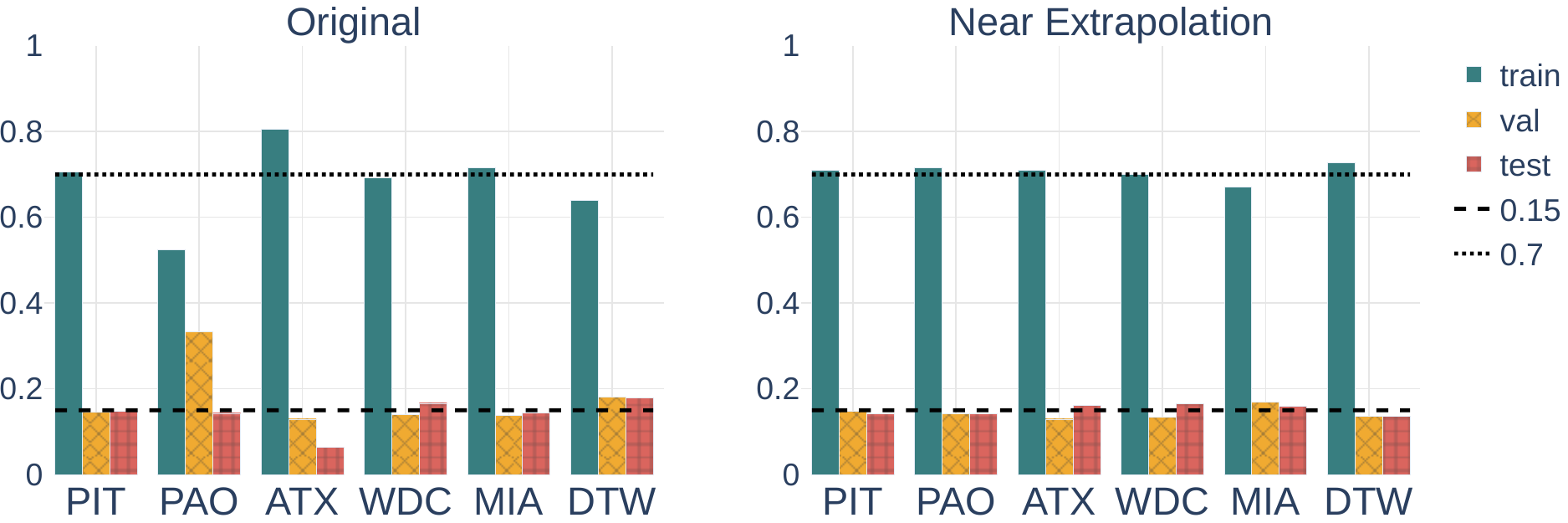}
    \caption{Inter-set city distribution for the Original and Geographically split data of Argoverse 2. The dotted and dashed lines represent the $70$\% and $15$\% target ratios respectively.}
    \label{fig:argo_city}
\end{figure}

\begin{figure*}[h]
  \centering
   \includegraphics[width=\linewidth, trim={90mm, 0mm, 90mm, 0mm}, clip]{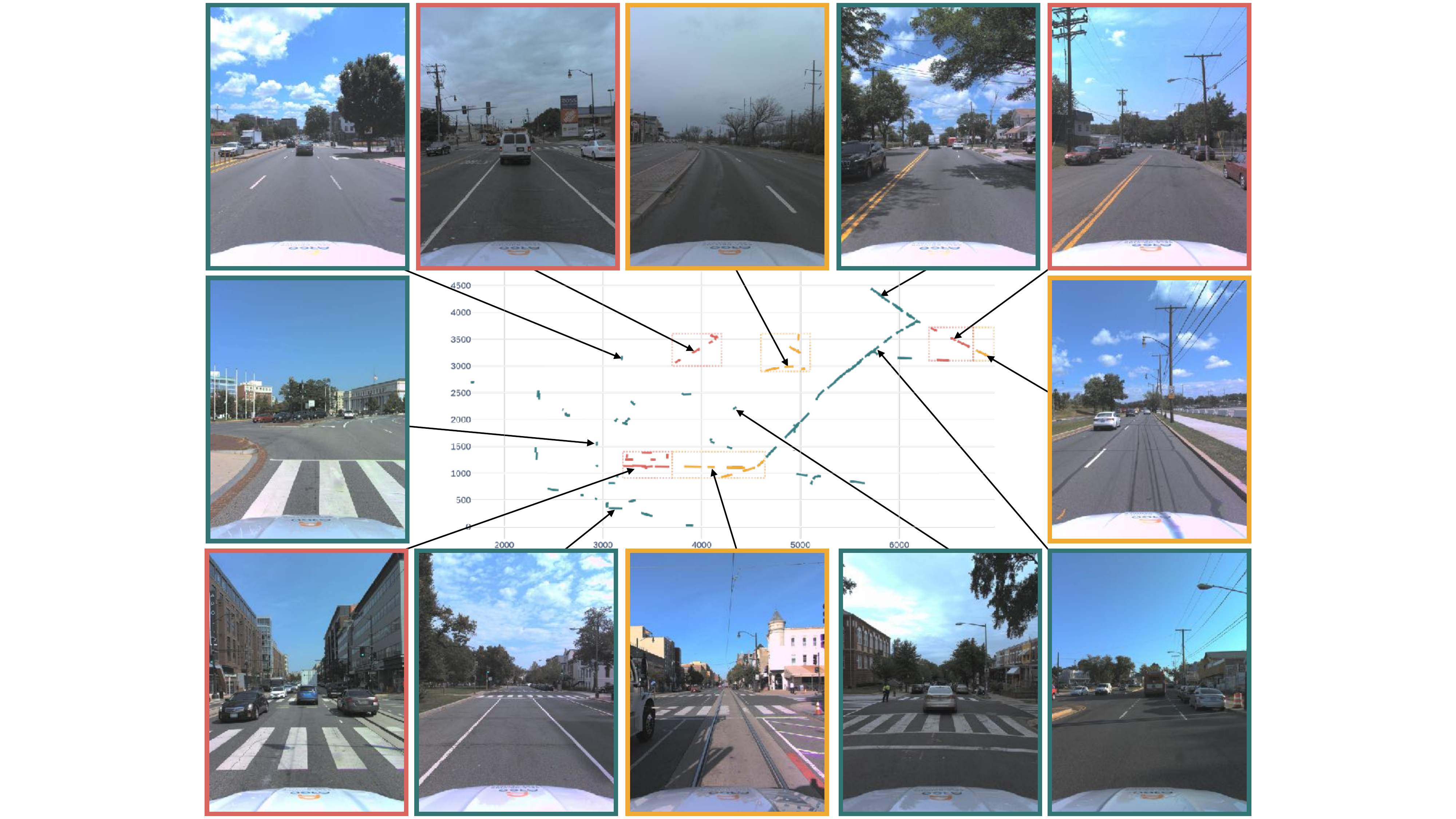}
   \caption{Samples from the Washington DC map in Argoverse 2 dataset, with marked training (green), validation (blue), and test (red) pose according to the Near Extrapolation split. Dotted polygons mark the boundaries of the validation and test zones. To enhance diversity in zone types within each set, regions from different parts of the city are incorporated. The downtown area in the southwest has different road characteristics from the sub-urban areas in the north and eastern parts.}
   \label{fig:argo-washington-thumbnails}
\end{figure*}

\begin{figure*}[h]
    \centering
    \begin{subfigure}[b]{0.45\textwidth}
        \centering
        \includegraphics[width=\linewidth, trim={0mm, 0mm, 30mm, 0mm}, clip]{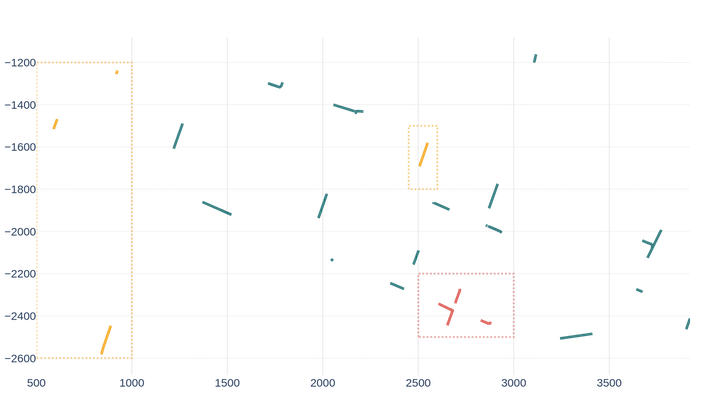}
        \caption{Austin}
    \end{subfigure}
    \begin{subfigure}[b]{0.45\textwidth}
    \centering
        \includegraphics[width=\linewidth, trim={0mm, 0mm, 30mm, 0mm}, clip]{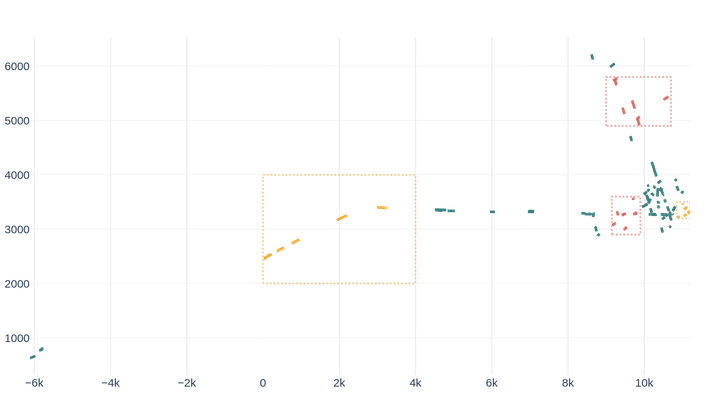}
        \caption{Detroit}
    \end{subfigure}
    
    \begin{subfigure}[b]{0.45\textwidth}
        \centering
        \includegraphics[width=\linewidth, trim={0mm, 0mm, 30mm, 0mm}, clip]{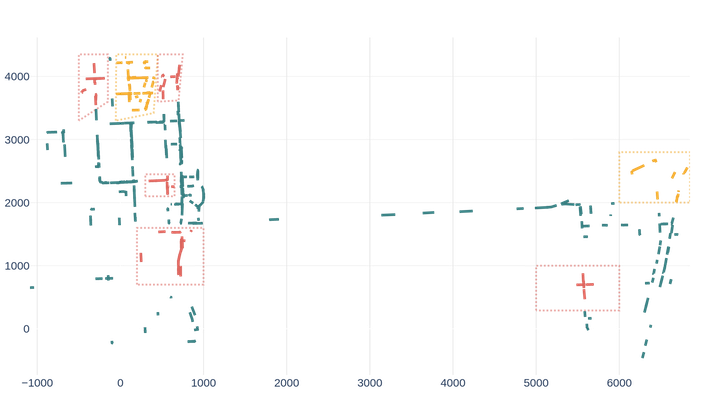}
        \caption{Miami}
    \end{subfigure}
    \begin{subfigure}[b]{0.45\textwidth}
        \centering
        \includegraphics[width=\linewidth, trim={0mm, 0mm, 30mm, 0mm}, clip]{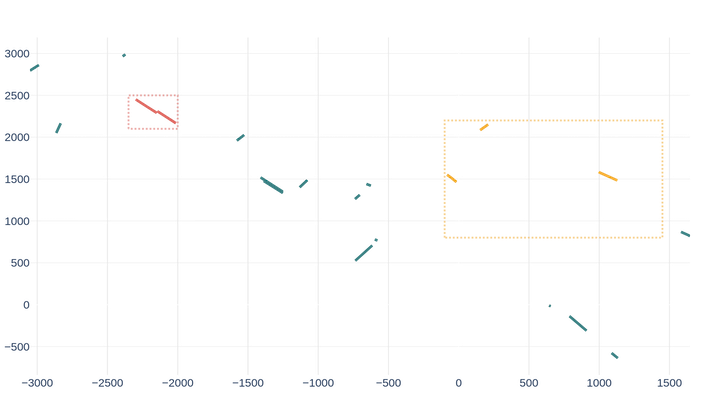}
        \caption{Palo Alto}
    \end{subfigure}

    \begin{subfigure}[b]{0.45\textwidth}
        \centering
        \includegraphics[width=\linewidth, trim={0mm, 0mm, 30mm, 0mm}, clip]{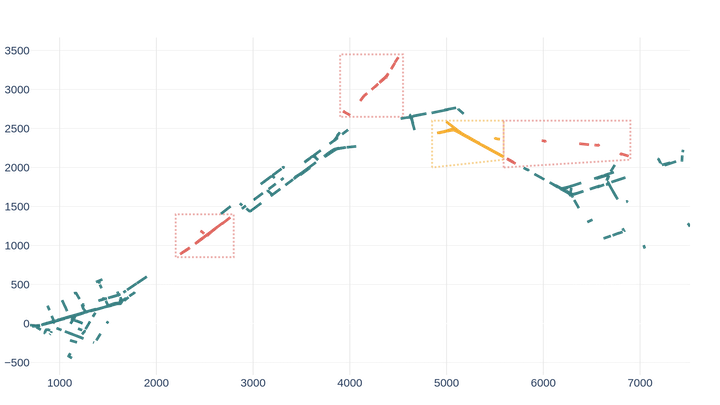}
        \caption{Pittsburgh}
    \end{subfigure}
    \begin{subfigure}[b]{0.45\textwidth}
        \centering
        \includegraphics[width=\linewidth, trim={0mm, 0mm, 30mm, 0mm}, clip]{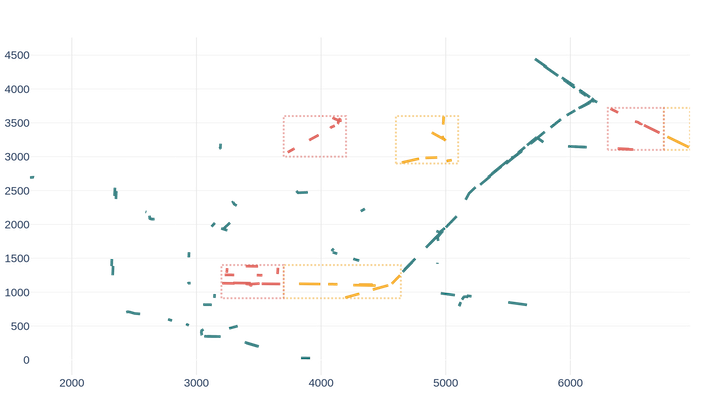}
        \caption{Washington DC}
    \end{subfigure}
   \caption{Near Extrapolation. Positions of samples in the nuScenes dataset, with the geographical areas of the validation and test sets outlined by dotted polygons. Training, validation, and test sets are distinguished by green, orange, and red colors, respectively. Regions from different parts of the cities are present in each set.}
   \label{fig:argo-sample-pos}
\end{figure*}

\begin{figure*}[h]
    \centering
    \begin{subfigure}[b]{0.45\textwidth}
        \centering
        \includegraphics[width=0.8\linewidth, trim={0mm, 0mm, 0mm, 18mm}, clip]{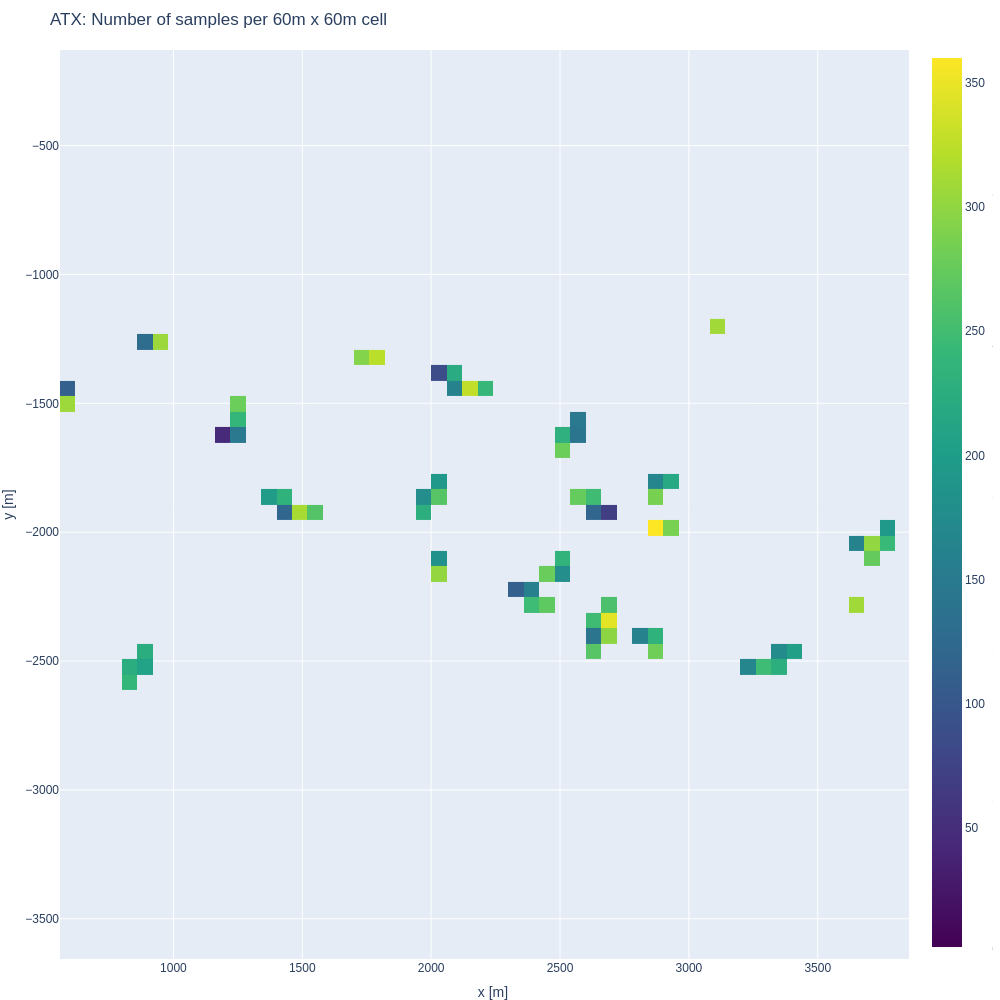}
        \caption{Austin}
    \end{subfigure}
    \begin{subfigure}[b]{0.45\textwidth}
    \centering
        \includegraphics[width=0.8\linewidth, trim={0mm, 0mm, 0mm, 18mm}, clip]{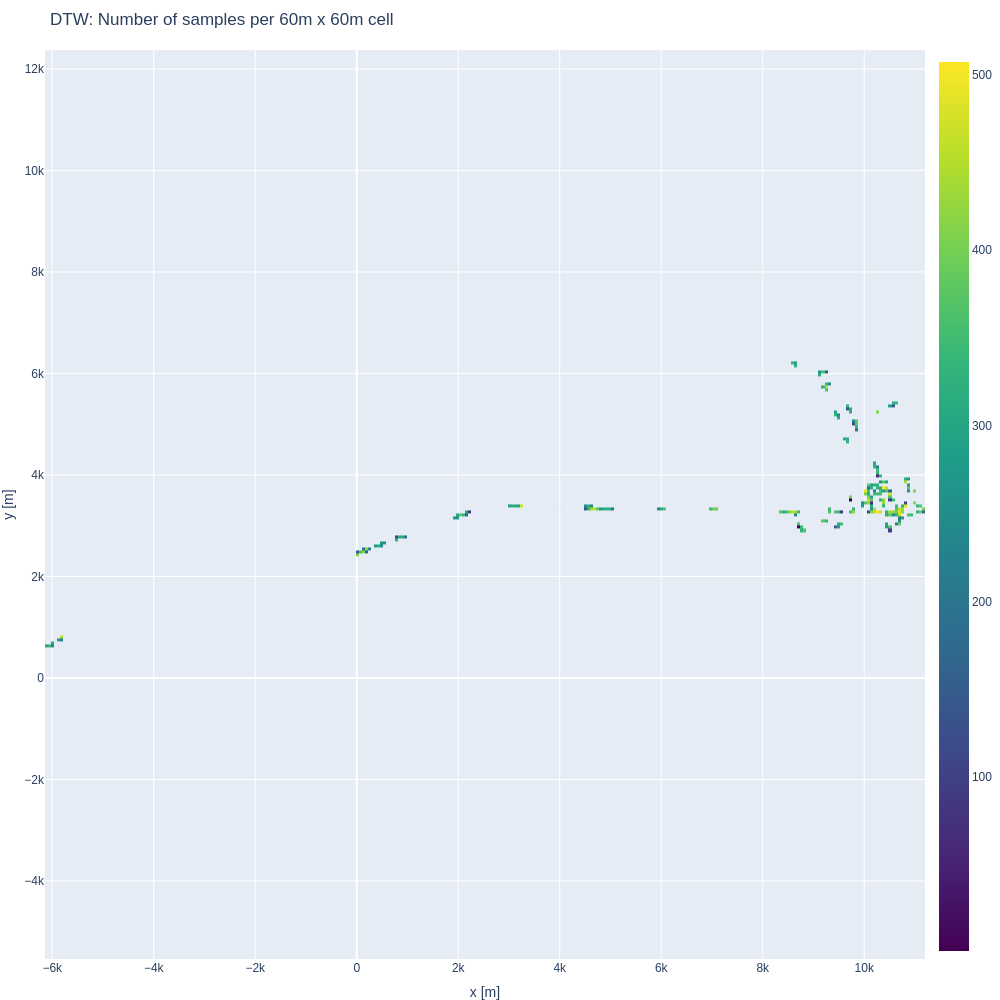}
        \caption{Detroit}
    \end{subfigure}
    
    \begin{subfigure}[b]{0.45\textwidth}
        \centering
        \includegraphics[width=0.8\linewidth, trim={0mm, 0mm, 0mm, 18mm}, clip]{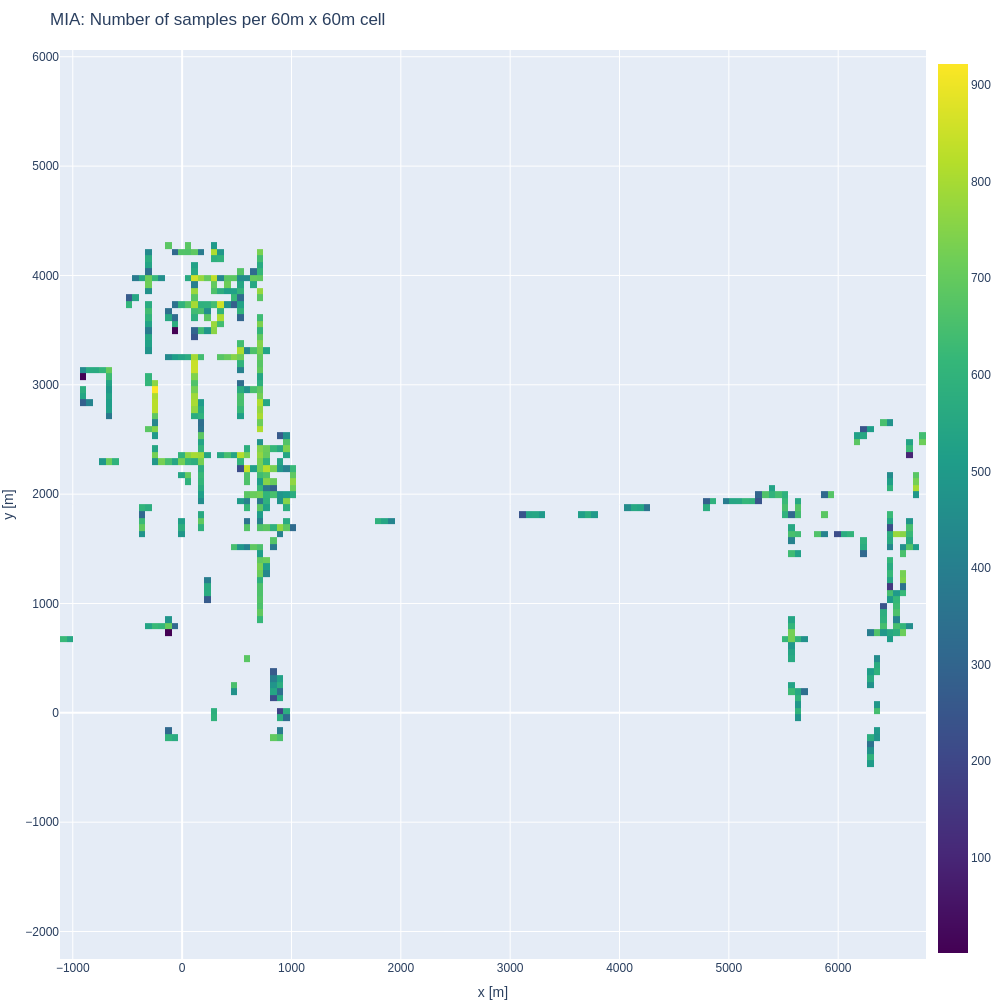}
        \caption{Miami}
    \end{subfigure}
    \begin{subfigure}[b]{0.45\textwidth}
        \centering
        \includegraphics[width=0.8\linewidth, trim={0mm, 0mm, 0mm, 18mm}, clip]{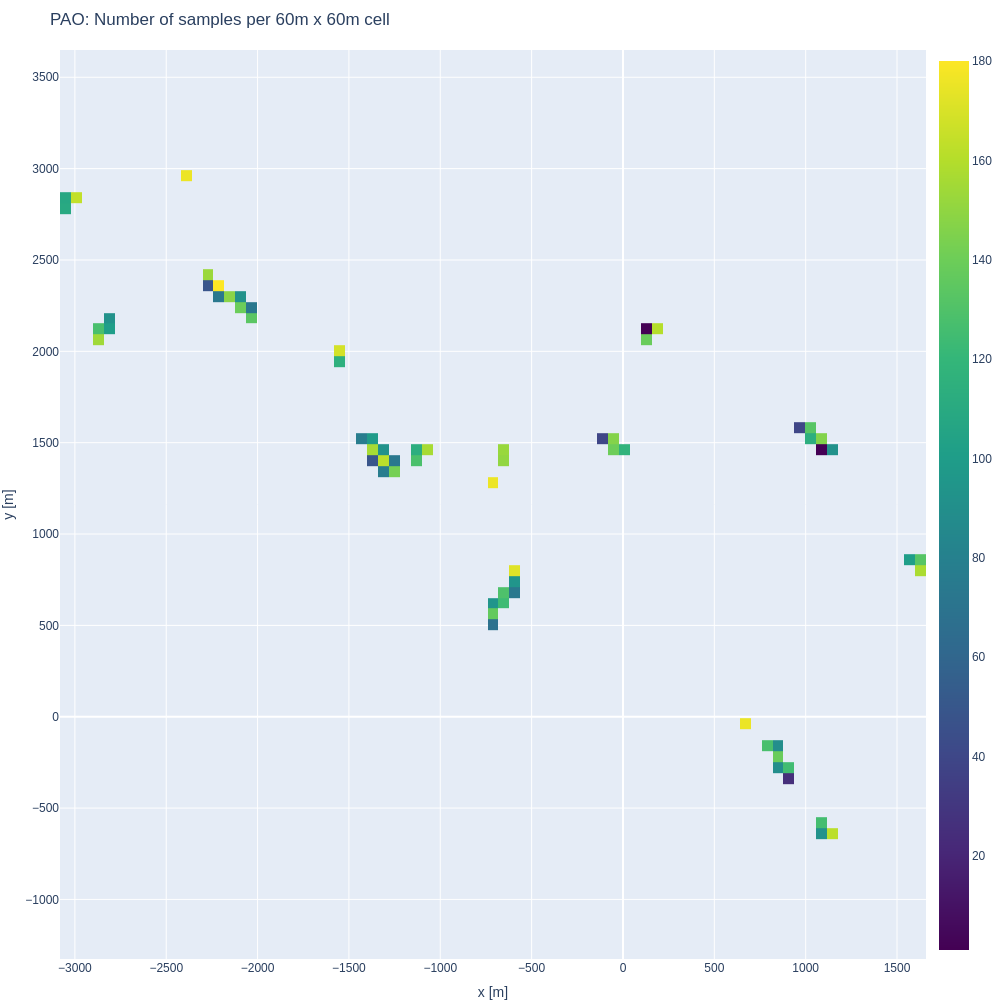}
        \caption{Palo Alto}
    \end{subfigure}

    \begin{subfigure}[b]{0.45\textwidth}
        \centering
        \includegraphics[width=0.8\linewidth, trim={0mm, 0mm, 0mm, 18mm}, clip]{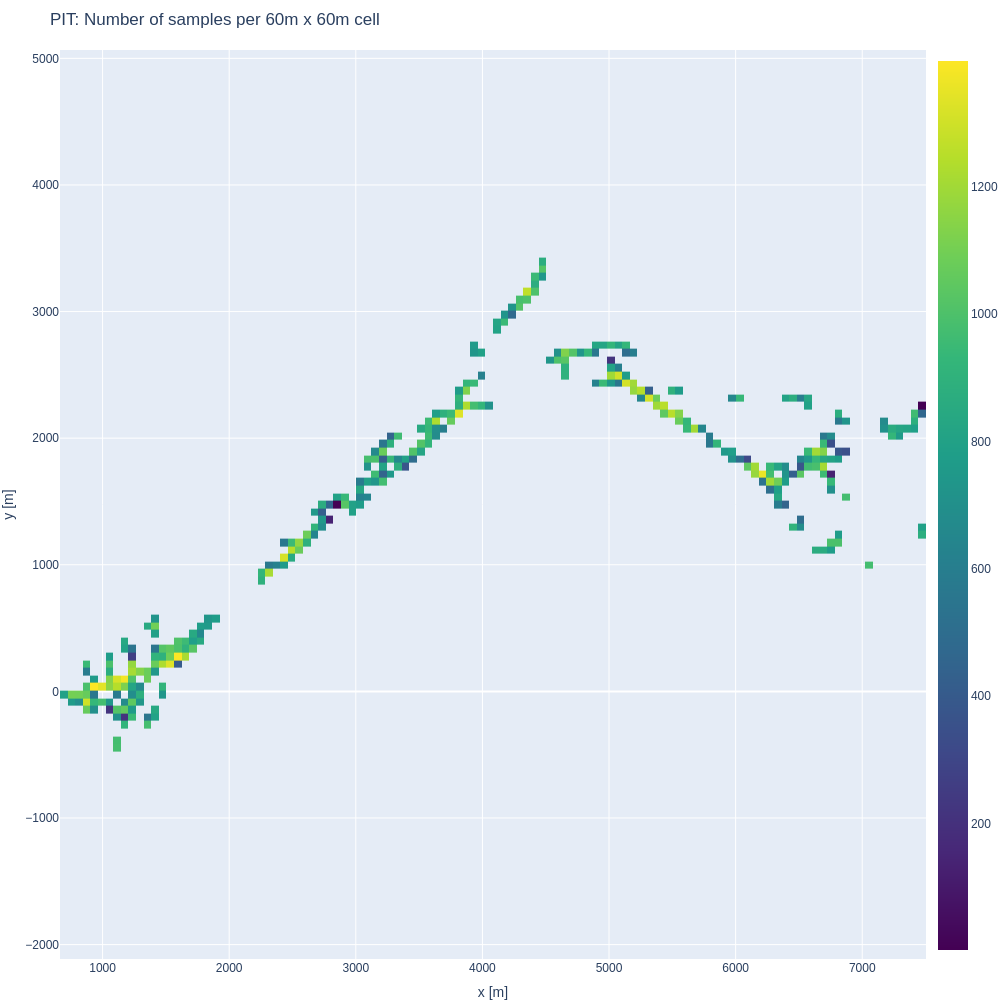}
        \caption{Pittsburgh}
    \end{subfigure}
    \begin{subfigure}[b]{0.45\textwidth}
        \centering
        \includegraphics[width=0.8\linewidth, trim={0mm, 0mm, 0mm, 18mm}, clip]{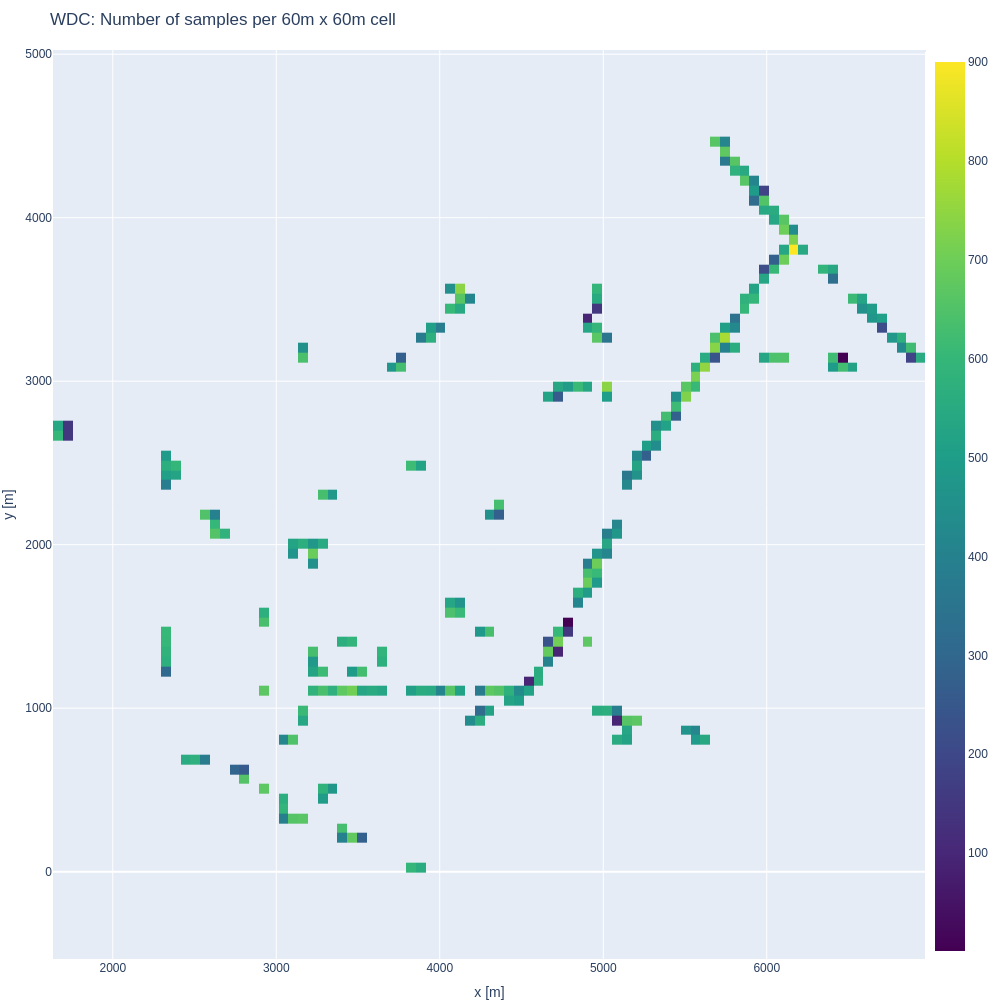}
        \caption{Washington DC}
    \end{subfigure}
   \caption{Heatmaps for the number of samples within 60m cells for Argoverse 2 dataset. Many cells contain a lot of samples, with the maximum number of samples in a single cell being 1398.}
   \label{fig:argo-heatmaps}
\end{figure*}

\paragraph{Argoverse 2}
As discussed in \cref{sec:new-splits-argo} it is possible to split Argoverse 2 on a sequence level while preserving zone class diversity. \cref{fig:argo_city} illustrates the distribution of the number of samples in each city. \cref{fig:argo-washington-thumbnails} further highlights the significance of diverse city areas in all sets by presenting a collection of images from various locations in Washington DC. The downtown area in the southwest exhibits different road characteristics from the sub-urban areas in the north and east. 
\cref{fig:argo-sample-pos} illustrates how the complete set of city maps are split, ensuring a diverse selection of areas in each set.
Heatmaps in \cref{fig:argo-heatmaps} represent the distribution of samples within each 60m cell.

\section{Extended Experiments}
\label{sec:extended-experiments}
To further investigate the effects of data splits and the amount of data, we perform additional experiments.

\paragraph{Far Extrapolation} 
We also train and evaluate the segmentation-based methods with city-wise folds. \cref{tab:citywise-segmentation} reports the performance for the folds in the Far Extrapolation split and their cross-validation mean. 
The performance on these folds are, similarly to the Near Extrapolation splits, lower than on the Original splits.
For nuScenes the performance on Near Extrapolation splits is already low, and the performance on the Far Extrapolation folds are on par. For Argoverse 2 the city-wise folds are performing worse than the Near Extrapolation split's validation set, but similar to the test set. This results in the cross-validation mean being consistently lower than the mean of the validation and test performance of the Near Extrapolation splits.
\begin{table}
\centering
\renewcommand{\arraystretch}{0.8}
\setlength{\tabcolsep}{2pt} 
\scalebox{0.7}{%
\begin{tabular}{l cc c c c c c}
\hline 
\toprule 
\multirow{2}{*}{\parbox[t]{2mm}{\multirow{2}{*}{\rotatebox[origin=c]{90}{}}}} & Model & Split & \lanemarkers & \roadedges & \zebras & Mean & CV \\ \cmidrule(lr){1-8} 
\multirow{4}{*}{\parbox[t]{2mm}{\multirow{4}{*}{\rotatebox[origin=c]{90}{nuScenes}}}}& 
\multirow{2}{*}{GKT} & A            & $10.6$ & $14.2$ & $0.8$  & $8.5$  & \multirow{2}{*}{$9.9$} \\ 
& & B                               & $14.7$ & $17.2$ & $1.6$  & $11.2$ & \\ \cmidrule{3-8} 
& \multirow{2}{*}{CVT} & A          & $13.1$ & $14.1$ & $2.2$  & $9.8$  & \multirow{2}{*}{$10.7$}\\ 
& & B                               & $14.8$ & $17.5$ & $2.6$  & $11.6$ & \\ \cmidrule{3-8} 
& \multirow{2}{*}{IPM} & A          & $28.1$ & $34.0$ & $12.1$ & $24.7$ & \multirow{2}{*}{$26.6$}\\ 
& & B                               & $33.6$ & $38.8$ & $13.0$ & $28.5$ & \\ \cmidrule{3-8} 
& \multirow{2}{*}{HDMapNet} & A     & $20.1$ & $20.7$ & $7.2$ & $16.0$ & \multirow{2}{*}{$27.3$}\\ 
& & B                               & $24.2$ & $24.4$ & $6.9$ & $18.5$ & \\ \cmidrule{1-8} 
\multirow{5}{*}{\parbox[t]{2mm}{\multirow{6}{*}{\rotatebox[origin=c]{90}{Argoverse 2}}}}
& \multirow{3}{*}{GKT} & A          & $28.2$ & $21.8$ & $7.1$  & $19.0$  & \multirow{3}{*}{$20.1$}   \\ 
& &                               B & $30.8$ & $25.8$ & $6.8$  & $21.1$ \\ 
& &                               C & $29.5$ & $23.7$ & $6.9$  & $20.0$ \\ \cmidrule{3-8} 
& \multirow{3}{*}{CVT} & A          & $29.0$ & $21.9$ & $9.5$  & $20.1$  & \multirow{3}{*}{$20.8$}\\ 
& & B                               & $31.9$ & $23.0$ & $7.6$  & $20.8$ \\
& & C                               & $30.6$ & $24.0$ & $9.3$  & $21.3$ \\ \cmidrule{3-8} 
& \multirow{3}{*}{IPM} & A          & $43.0$ & $38.2$ & $24.6$ & $35.3$ & \multirow{3}{*}{$37.4$}\\ 
& & B                               & $48.6$ & $43.3$ & $25.8$ & $39.2$ \\
& & C                               & $45.1$ & $41.6$ & $26.8$ & $37.8$ \\ 
\bottomrule
\end{tabular}}
\caption{Segmentation-based methods' IoU on the city-wise folds of the Far Extrapolation split and their corresponding cross-validation mean (CV).}
\label{tab:citywise-segmentation}
\end{table}

\paragraph{Training set extension}
To explore how the amount of data affects the performance we extend the training set with the validation samples, effectively increasing the training set with $20$\%. \cref{tab:val-to-train-performance} shows a boost in the test performance for both segmentation- and vector-based methods. 
The impact is greater on nuScenes, but Argoverse 2 also benefits from the added data, indicating that more extensive datasets are necessary for learning online mapping. For instance, the extra data has a higher impact for MapTR on Argoverse 2, $+1.4$ mAP, than the choice of lifting method, $+0.7$ mAP. 
On nuScenes, the impact is greater, but also similarly large as using LSS in comparison to GKT for lifting, $+1.9$ and $+2.0$ respectively. The lifting methods are further discussed in \cref{par:lifting-method} and shown in \cref{tab:maptr-lifter-ablation}.
\def\aoldlm{$15.3$}
\def\aoldre{$17.3$}
\def\aoldpc{$9.0$}
\def\aoldmap{$13.9$}

\def\boldlm{$24.4$}
\def\boldre{$26.3$}
\def\boldpc{$14.7$}
\def\boldmap{$21.8$}

\def\coldlm{$25.3$}
\def\coldre{$26.2$}
\def\coldpc{$14.9$}
\def\coldmap{$22.2$}

\def\doldlm{$17.3$}
\def\doldre{$21.6$}
\def\doldpc{$15.7$}
\def\doldmap{$18.2$}

\def\eoldlm{$18.8$}
\def\eoldre{$25.3$}
\def\eoldpc{$17.6$}
\def\eoldmap{$20.6$}

\def\foldlm{$20.5$}
\def\foldre{$4.6$}
\def\foldpc{$17.5$}
\def\foldmap{$20.9$}

\def\goldlm{$23.4$}
\def\goldre{$40.5$}
\def\goldpc{$14.8$}
\def\goldmap{$26.2$}

\def\holdlm{$25.3$}
\def\holdre{$42.1$}
\def\holdpc{$18.6$}
\def\holdmap{$28.7$}

\def\ioldlm{$26.7$}
\def\ioldre{$40.3$}
\def\ioldpc{$20.1$}
\def\ioldmap{$29.1$}

\def\joldlm{$19.9$}
\def\joldre{$33.3$}
\def\joldpc{$5.9$}
\def\joldmap{$19.7$}

\def\koldlm{$21.9$}
\def\koldre{$36.1$}
\def\koldpc{$6.8$}
\def\koldmap{$21.6$}

\begin{table}
\centering
\renewcommand{\arraystretch}{0.8} 
\setlength{\tabcolsep}{2pt} 
\scalebox{0.8}{%
\begin{tabular}{l cc c c c c}
\hline 
\toprule 
\multirow{2}{*}{\parbox[t]{2mm}{\multirow{2}{*}{\rotatebox[origin=c]{90}{}}}} & Model & Split & \lanemarkers & \roadedges & \zebras & Mean \\ \cmidrule(lr){1-7} 
\multirow{6}{*}{\parbox[t]{2mm}{\multirow{4}{*}{\rotatebox[origin=c]{90}{nuScenes}}}}& \multirow{2}{*}{HDMapNet} & \oursplitname               & \aoldlm & \aoldre & \aoldpc & \aoldmap \\ 
& & \oursplitname$\cup$                  & \boldlm & \boldre & \boldpc & \boldmap \\ \cmidrule{3-7} 
& \multirow{2}{*}{VectorMapNet} & \oursplitname           & \doldlm & \doldre & \doldpc & \doldmap \\ 
& & \oursplitname$\cup$                                  & \eoldlm & \eoldre & \eoldpc & \eoldmap \\ \cmidrule{3-7} 
& \multirow{2}{*}{MapTR} & \oursplitname           & \joldlm & \joldre & \joldpc & \joldmap \\ 
& & \oursplitname$\cup$                            & \koldlm & \koldre & \koldpc & \koldmap \\ \cmidrule{3-7} 
& \multirow{2}{*}{MapTRv2} & \oursplitname                & \goldlm & \goldre & \goldpc & \goldmap \\ 
& & \oursplitname $\cup$                                     & \holdlm & \holdre & \holdpc & \holdmap \\ \cmidrule{1-7} 
\multirow{4}{*}{\parbox[t]{2mm}{\multirow{4}{*}{\rotatebox[origin=c]{90}{Argoverse 2}}}}
& \multirow{2}{*}{VectorMapNet} & \oursplitname       & $35.0$ & $32.4$ & $31.3$ & $32.9$     \\ 
& &                               \oursplitname$\cup$ & $37.5$ & $33.1$	& $32.7$ & $34.4$ \\ \cmidrule{3-7} 
& \multirow{2}{*}{MapTR} & \oursplitname & $45.2$	&$48.3$&	$50.9$&	$48.2$  \\ 
& & \oursplitname$\cup$ & $47.3$	&$49.4$&	$52.0$&	$49.6$ \\ \cmidrule{3-7} 
& MapTRv2 & \oursplitname &  $56.6$ & $53.5$ & $55.6$ & $55.2$\\ 
& 2D & \oursplitname $\cup$  &  $59.5$ &	$54.7$ &	$53.4$ & $55.9$ \\
\bottomrule
\end{tabular}}
\caption{Increasing training data by $15$\% using the union of training and validation samples, marked by $\cup$, improves test performance. IoU for HDMapNet and mAP for the other methods.}
\label{tab:val-to-train-performance}
\end{table}

\paragraph{Hyperparameter-search}
For MapTRv2 on the Near Extrapolation split on nuScenes, we investigate various hyperparameters related to overfitting on the training set, \ie, weight decay, learning rate, and training epochs. 
Interestingly, we can in \cref{tab:hyper-search} observe only minor differences and the parameters initially employed for training on the Original split seem equally effective for the geographically disjoint split. 

\begin{table}
\centering
\setlength{\tabcolsep}{2pt} 
\scalebox{0.8}{%
\begin{tabular}{l c cc}
\hline 
\toprule 
&&                       \multicolumn{2}{c}{LR}              \\ 
\multirow{3}{*}{\parbox[t]{2mm}{\multirow{4}{*}{\rotatebox[origin=c]{90}{WD}}}} &  & $6e^{-4}$ & $1e^{-4}$ \\ \cmidrule(lr){3-4} 
                        & $0.05$              & $27.0$    & $26.5$ \\
                        & $0.10$              & $26.7$    & $27.2$ \\
                        & $0.15$              & $27.1$    & $26.5$ \\
\bottomrule
\end{tabular}}
\caption{MapTRv2 show robustness to different hyperparameters, learning rate (LR) and weight decay (WD) on nuScenes Near Extrapolation split.}
\label{tab:hyper-search}
\end{table}

\section{Qualitative Results}
\label{sec:qualitative-results}
\begin{figure*}[h]
  \centering
   \includegraphics[width=\linewidth, trim={25mm, 0mm, 25mm, 0mm}, clip]{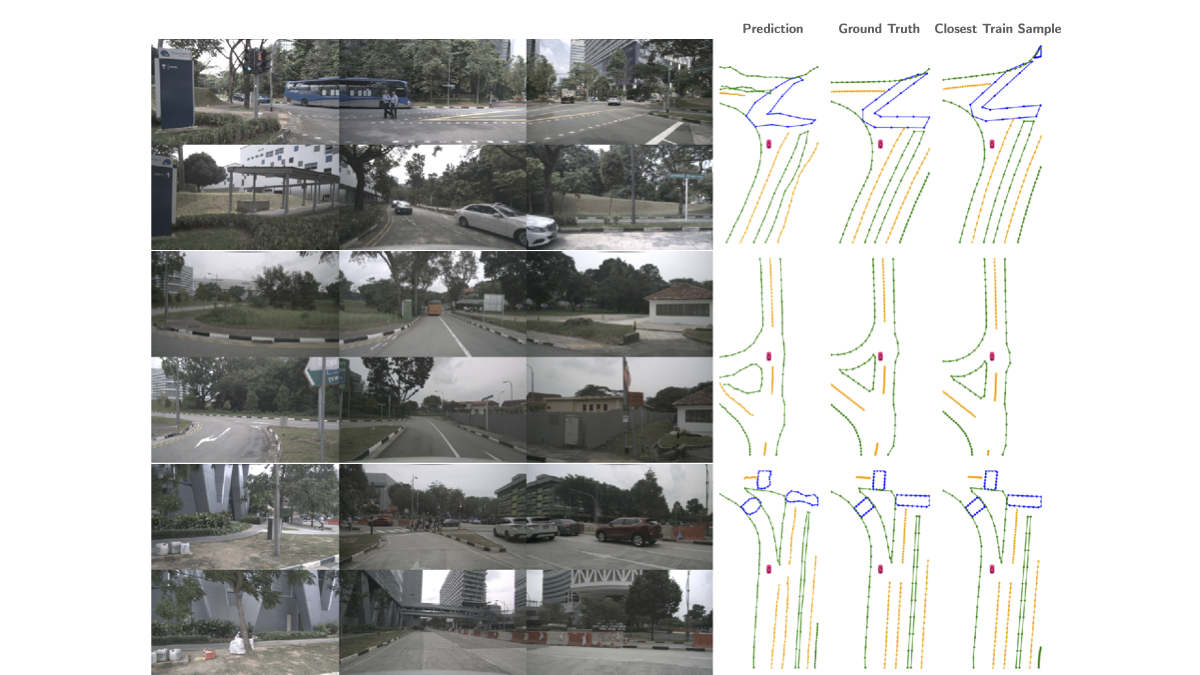}
   \caption{Multiple examples of validation or test prediction from MapTR on nuScenes, corresponding ground truth, and the closest training sample's ground truth. The close similarities between the closest training samples and the evaluation samples are evident in each example.}
   \label{fig:nusc-qual-example-supmat-2}
\end{figure*}

\paragraph{nuScenes}
\cref{fig:nusc-qual-example-supmat-2} portrays three examples with input images, the evaluation prediction, its ground truth, and the closest training sample. Despite not being captured from the exact same pose, these instances demonstrate striking similarities between the evaluation and closest training pose. This underscores that the method, having encountered the closest training sample during training, can achieve accurate predictions through memorization and retrieval of these examples at test time. Additional examples can be seen in the videos to be part of the project webpage.

\begin{figure*}[h]
  \centering
   \includegraphics[width=\linewidth, trim={0mm, 0mm, 0mm, 0mm}, clip]{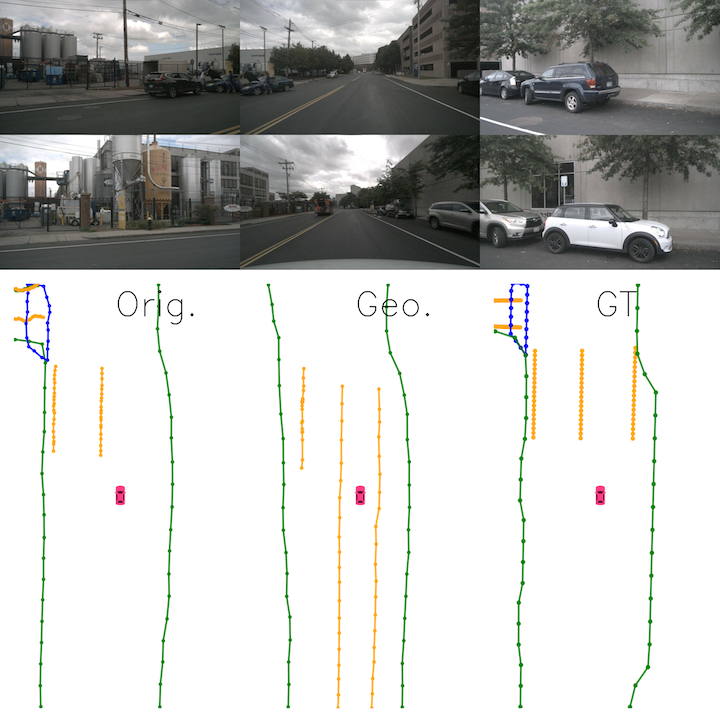}
   \caption{nuScenes test prediction from MapTR trained on Original (Orig.) and Geographically disjoint (Geo.), here Near Extrapolation, along with the ground truth (GT). Dividers, Boundaries, and Pedestrian crossings are visualized in orange, green, and blue respectively. Despite occlusion on the left side by opposing lane vehicles, the method trained on the original split accurately predicts them. In contrast, the model trained on geographically disjoint splits fails to detect them. On the other hand, the model trained on geographically disjoint split data successfully identifies dividers near the ego vehicle, even though they are absent in the ground truth.}
   \label{fig:nusc-qual-example-supmat-3}
\end{figure*}

\cref{fig:nusc-qual-example-supmat-3} compares the predictions of a sample included in the test set of both the Original and Near Extrapolation splits. It demonstrates that a model trained on the Original split can predict dividers, boundaries, and pedestrian crossings occluded by vehicles in the opposing lane accurately. Thus making it tempting to speculate that the method has memorized this information.
Furthermore, it shows that the model trained on geographically disjoint data only identifies the dividers near the ego vehicle. These dividers are visible in the images, but absent in the ground truth, indicating that the model has learned to generalize better.

\begin{figure*}[h]
  \centering
   \includegraphics[width=\linewidth, trim={140mm, 0mm, 140mm, 10mm}, clip]{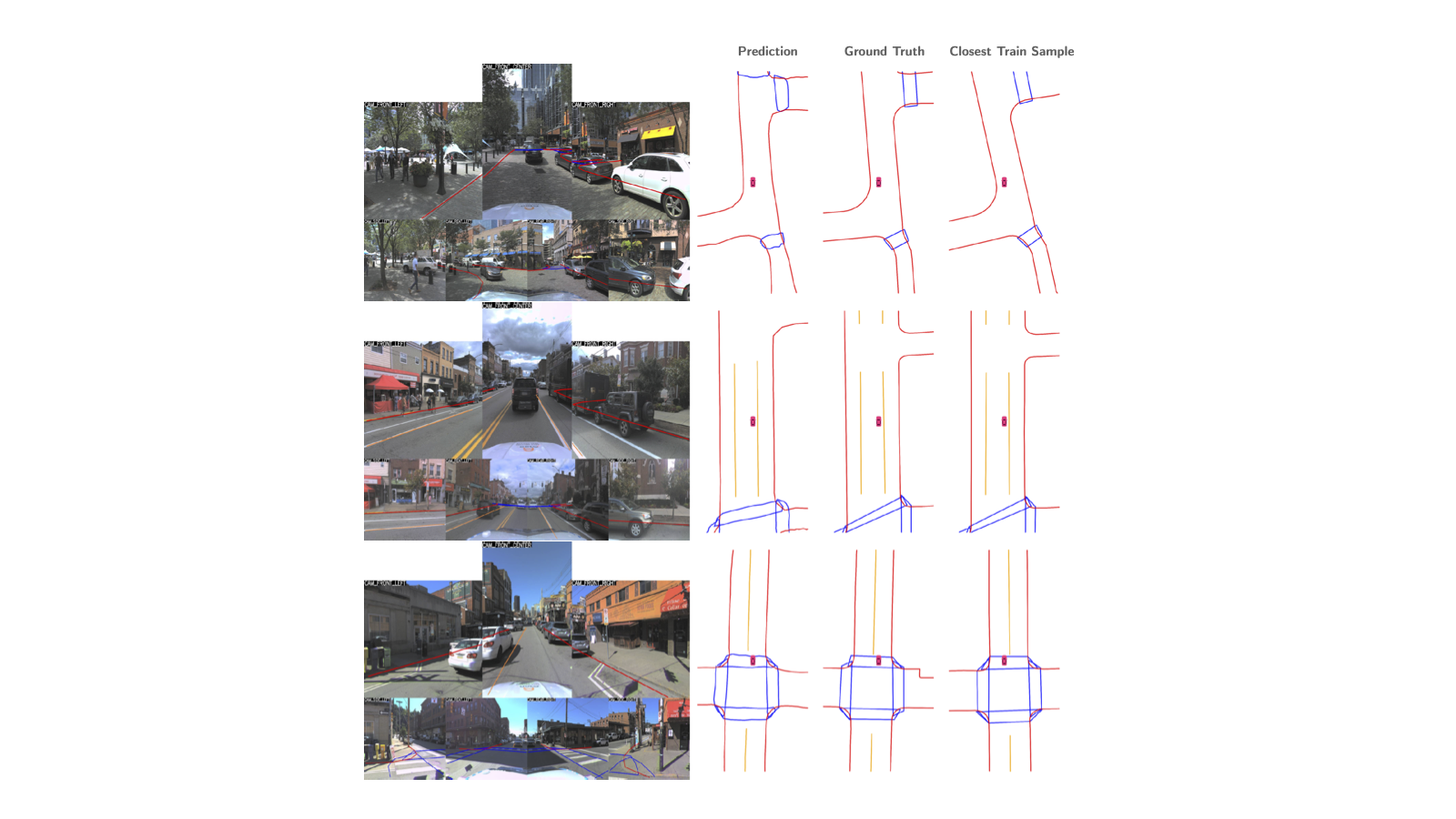}
   \caption{Multiple instances of validation or test predictions from MapTRv2 on Argoverse 2, alongside corresponding ground truth and the ground truth of the nearest training sample. The close similarities between the nearest training samples and the evaluation samples are apparent in each example. In the top illustration, the closest training sample exhibits a slight rotation, but the positions are very similar. In the bottom example, the closest training sample is from the lane adjacent to the evaluation sample}
   \label{fig:argo-qual-example-supmat-1}
\end{figure*}

\paragraph{Argoverse 2}
In \cref{fig:argo-qual-example-supmat-1}, we present three examples featuring input images, the evaluation prediction, its ground truth, and the closest training sample. While not being from the exact same pose, \eg in the top example the closest training pose is slightly rotated, and in the bottom from an adjacent lane, it is still plausible for a method to achieve a high score on the test sample by memorizing the map and images from the training sample, and then recall and slightly shift and rotate that map at test time. Additional examples will be available on the project webpage.

\begin{figure*}[h]
  \centering
   \includegraphics[width=\linewidth, trim={30mm, 0mm, 30mm, 0mm}, clip]{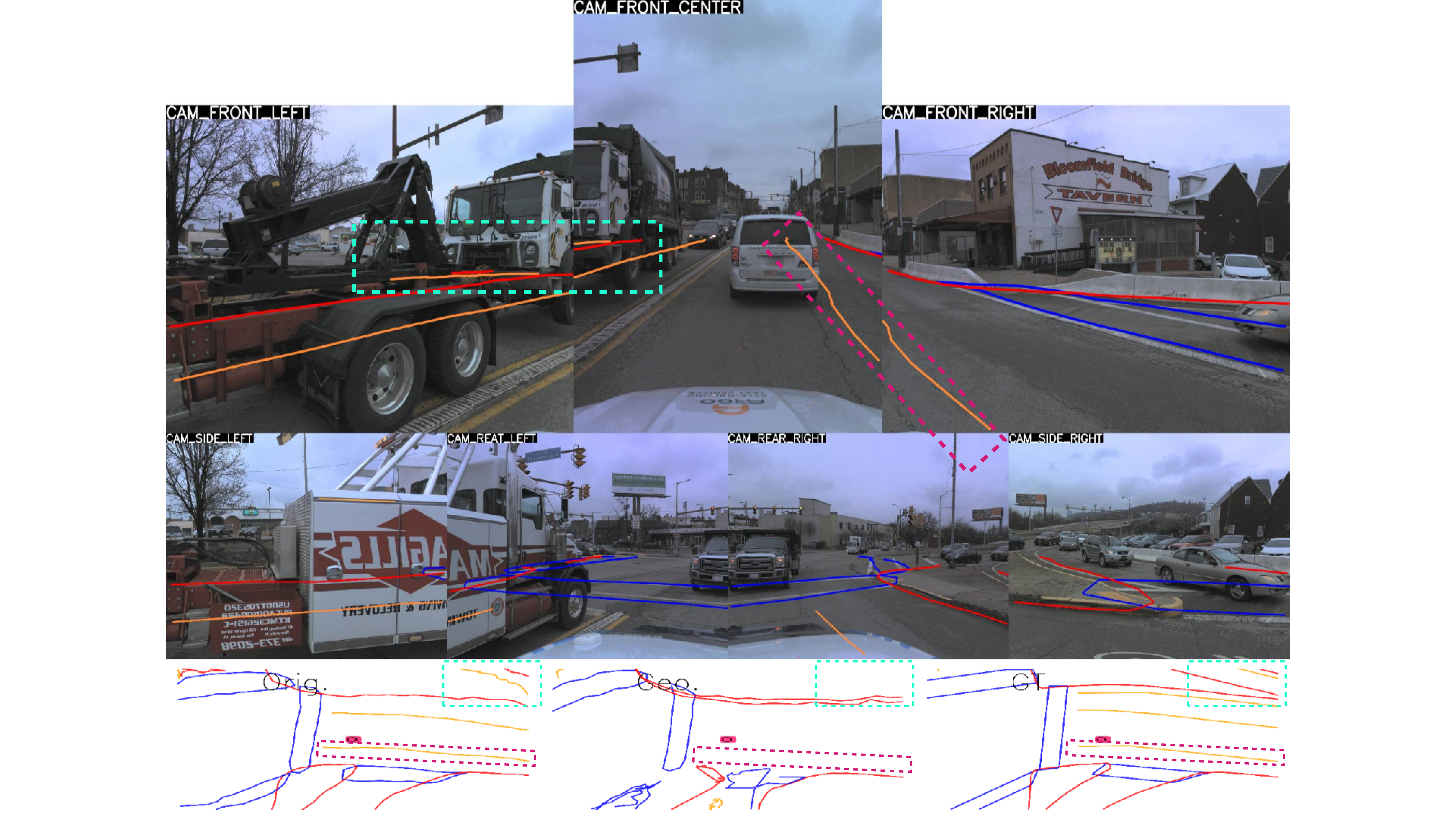}
   \caption{Argoverse 2 test prediction by mapTRv2 trained on Original (Orig.) and Geographical (Geo.), here Near Extrapolation, splits along with the ground truth (GT). Dividers, Boundaries, and Pedestrian crossings are visualized in orange, red, and blue, respectively. The predictions in the image view are from training on the Original split. Here, the predictions behind the truck on the left side, most notably the divider and boundary highlighted with the teal box, ought to be difficult to predict. Additionally, the model effectively predicts the lane divider to the right of the ego vehicle, highlighted by the pink box, even though there is no visible lane divider present in the image. It is worth noting that this may not solely be due to memorization, as the model could learn, \eg, consistent data annotations and hints from road dividers and road width to accurately predict this non-visible lane divider. }
   \label{fig:argo-qual-compare}
\end{figure*}

\cref{fig:argo-qual-compare} illustrate comparisons between predictions derived from a sample included in both the Original and Geographically disjoint splits' test set, along with the ground truth. Despite the inherent difficulty in predicting objects situated behind a truck on the left side, the model trained on the original split demonstrates commendable accuracy in its estimations.
The model also effectively predicts the lane divider to the right of the ego vehicle, when not visibly present in the image but existing in the ground truth. It is worth noting that this may not be due to memorization, as the model could learn, e.g., consistent data annotations and hints from road dividers and road width to accurately predict this non-visible lane divider.

\newpage

\end{document}